\documentclass[times,10pt]{article}
\usepackage{xcolor}
\usepackage{graphicx}
\usepackage{subcaption} 
\usepackage[sort]{natbib}
\usepackage{multirow}
\usepackage[utf8]{inputenc}
\usepackage{authblk}
\usepackage{tabularx}
\usepackage{booktabs}
\usepackage{hyperref}
\usepackage[margin=1cm]{caption} 
\usepackage{amsmath}
\usepackage{amssymb}
\usepackage{amsfonts}
\usepackage{float}
\usepackage{bm}
\usepackage{subcaption}
\usepackage{lipsum} 
\usepackage[includefoot]{geometry}
\title{\bf Evaluating the effects of Data Sparsity on the Link-level Bicycling Volume Estimation: A Graph Convolutional Neural Network Approach}
\author[]{}

\author[a]{
    Mohit Gupta \thanks{Corresponding Author}
}

\author[a]{
    Debjit Bhowmick
}

\author[b]{
   Meead Saberi 
}
\author[c]{
   Shirui Pan 
} 

\author[a]{
    Ben Beck \thanks{
    Corresponding Author \\
    \hspace*{1cm}\textit{E-mail address:} \href{mailto: mohit.gupta1@monash.edu}{mohit.gupta1@monash.edu} (M. Gupta), \href{mailto: ben.beck@monash.edu} {ben.beck@monash.edu} (B. Beck)}
}
\tiny{\affil[a]{\scriptsize School of Public Health and Preventive Medicine, Monash University, Melbourne, Australia}
\affil[b]{School of Civil and Environmental Engineering, Research Centre for Integrated Transport Innovation (rCITI)}
\affil[c]{School of Information and Communication Technology, Griffith University, Brisbane, Australia}}
\date{}

\begin{document}
\maketitle
\begin{abstract}
Accurate bicycling volume estimation is crucial for making informed decisions and planning about future investments in bicycling infrastructure. 
However, traditional link-level volume estimation models are effective for motorised traffic but face significant challenges when applied to the bicycling context because of sparse data and the intricate nature of bicycling mobility patterns. 
To the best of our knowledge, we present the first study to utilize a Graph Convolutional Network (GCN) architecture to model link-level bicycling volumes and systematically investigate the impact of varying levels of data sparsity (0\%–99\%) on model performance, simulating real-world scenarios.
We have leveraged Strava Metro data as the primary source of bicycling counts across 15,933 road segments/links in the City of Melbourne, Australia. 
To evaluate the effectiveness of the GCN model, we benchmark it against traditional machine learning models, such as linear regression, support vector machines, and random forest. 
Our results show that the GCN model outperforms these traditional models in predicting Annual Average Daily Bicycle (AADB) counts, demonstrating its ability to capture the spatial dependencies inherent in bicycle traffic network. 
While GCN remains robust up to 80\% sparsity, its performance declines sharply beyond this threshold, highlighting the challenges of extreme data sparsity. These findings underscore the potential of GCNs in enhancing bicycling volume estimation, while also emphasizing the need for further research on methods to improve model resilience under high-sparsity conditions.  Our findings offer valuable insights for city planners aiming to improve bicycling infrastructure and promote sustainable transportation. 
\end{abstract}


\section{Introduction}

Bicycling has great potential to positively impact public and environmental health by being a sustainable and active mode of transportation. 
It is acknowledged as a viable means of reducing the external costs of motorised transportation, such as air and noise pollution \citep{jeong2022characteristics}, traffic congestion \citep{szell2018crowdsourced}, and the negative impacts of sedentary lifestyles on health. 
However, bicycling participation is low, injury rates are on the rise, and the primary barrier to increased bicycling is the perception of safety \citep{beck2017road, pearson2023adults, pearson2023barriers} and inadequate infrastructure \citep{utriainen2023review}.
Therefore, it is crucial to provide connected networks of high-quality infrastructure to enable more people to choose bicycling as their preferred mode of transportation. 
Accurate estimation of bicycling volume on individual street segments (link-level prediction) is essential for understanding bicycling demand across the network, providing critical insights to inform safety and for making informed decisions regarding infrastructure investments that will maximize impact \citep{bhowmick2023systematic}. 
By understanding the patterns and trends in bicycling activity across different areas of the city, planners can strategically allocate resources to improve existing infrastructure and develop new cycling facilities in areas with gaps in connectivity, in areas with high cycling demand, and in areas with poor safety.

In the broader transportation domain, motor vehicle volume modeling \citep{xu2020ge, lee2020emerging, lee2021strava, qiu2020topological} has evolved significantly over time, offering increasingly accurate predictions with notable benefits for various transportation applications  \citep{yu2021deep, trirat2023mg}. 
This progress can be attributed to advancements in data collection, computational techniques, and machine learning algorithms. 
Historical traffic data, real-time traffic sensor information, and spatial data have become more accessible, providing a richer foundation for model development. 
Machine learning algorithms, and deep learning techniques in particular, have emerged as powerful tools for link-level volume estimation, enabling models to learn non-linear patterns and adapt to changing traffic conditions.

While link-level volume estimation models have proven effective in the motorised traffic domain, their application to bicycling presents unique challenges and opportunities. 
Bicycling volume estimation studies have been challenged by data sparsity, limited spatial and temporal coverage, and the low granularity of the available count data \citep{bhowmick2023systematic}. 
Among these challenges, data sparsity emerges as a formidable obstacle, restricting our ability to comprehensively understand and estimate bicycle traffic patterns. 
In the context of bicycling volume prediction, data sparsity refers to the presence of gaps or missing bicycling count data in the dataset that represents the volume of bicycling activities over a certain period of time or in a specific area.
Bicycle traffic data \citep{saberi2024modeling} is inherently more challenging to collect compared to motorised vehicle traffic data, which amplifies the complexities associated with data sparsity \citep{broach2012cyclists, lee2020emerging}. Unlike motorised vehicles that utilize well-established road networks equipped with sensors and traffic cameras, bicyclists often traverse a wider range of infrastructure, including sidewalks, bike lanes, and shared-use paths. Additionally, many of the existing infrastructure to capture motorised vehicle traffic data cannot capture bicycle traffic data \citep{kothuri2017bicycle, olmos2020data}.
Therefore, the absence of equivalent count data and relevant infrastructure data for bicycling has severely limited our ability to progress bicycling volume prediction.

The evolution of transportation modeling algorithms has progressed from statistical methods to advanced machine learning models, ultimately leading to the emergence of deep learning approaches \citep{boukerche2020machine, haghighat2020applications}. 
Early transportation modeling relied on statistical methods such as the Kalman filter, ARIMA, and variants due to their sound mathematical foundations. 
However, such approaches often resulted in poor performance as these models work on the assumption of linearity and stationarity in the data, which does not hold true for the traffic domain \citep{jiang2022graph}. 
Machine learning models such as Support Vector Machine (SVM) \citep{fu2016vehicle} and K-Nearest Neighbor (KNN) \citep{may2008vector} can model non-linearity and complex relationships in the data. 
However, the core assumption of independent and identically distributed data (IID) in machine learning algorithms does not hold true for road network data because each instance is linked to and thus dependent on neighboring road segments. For instance, traffic flow on one road segment is influenced by the conditions of adjacent segments, especially where they intersect.  
Therefore, their performance drops significantly in big data contexts such as transportation modeling \citep{tedjopurnomo2020survey}.
Recently, the uptake of deep learning models \citep{liu2020urban} such as Convolutional Neural Networks (CNN) and Recurrent Neural Networks (RNN), in various domains including natural language processing and computer vision, has attracted applications in the transportation domain.
Researchers have used RNNs \citep{lv2018lc} to extract the temporal correlations and CNNs \citep{ma2017learning} to capture the spatial correlations in the motorised road network data by representing it as a grid-based structure. 
Grid-based structures refer to representing the data features in the 2-dimensional space where each node is connected to its neighboring nodes. 
However, the traffic conditions can be influenced not only by nearby areas but also by areas that are spatially far apart but still connected within the network. This means that spatial dependencies might extend beyond just local areas. 
Graphs are therefore an essential tool to represent such spatial relationships effectively, as they capture the road network topology and are powerful in modeling these dependencies.

A graph is a data structure that consists of nodes (vertices) and edges (links). 
In a road network, nodes represent intersections, while edges represent the road segments that connect them.
Graph data is different from a fixed-sized grid as it has a variable number of unordered nodes, which makes it challenging for existing machine learning models to effectively process and analyze.
Traditional machine learning models are designed for structured data with fixed dimensions, making them less adaptable to the irregular and non-Euclidean nature of graph data.
Graph Neural Network (GNN) \citep{jiang2022graph} is a class of deep learning techniques designed to work with complex, graph-structured data and offers a simple method for performing node-level, edge-level, and graph-level prediction tasks. 
By leveraging the graph structure of road networks, GNNs can effectively capture the relationships between various elements in the road network such as connectivity and fine-grained spatial and temporal dependencies \citep{rahmani2023graph}. 

Although GNNs have demonstrated state-of-the-art performance in predicting motor vehicle volumes \citep{li2017diffusion, yu2017spatio, zhang2018gaan, zhao2019t}, they have not yet been applied to modeling bicycling volumes due to the limited availability of relevant data and associated challenges. 
This study aims to address this gap by developing a robust GCN architecture for link-level bicycling volume modeling in the City of Melbourne and evaluate the effects of data sparsity on model performance using Strava Metro Data \citep{lee2021strava}. 
To validate the performance of our proposed GCN model, we benchmark it against traditional machine learning models, including linear regression, support vector machines, and random forest. 
We systematically introduce varying degrees of sparsity into the bicycle network to explore the impact of sparsity on model performance. 
Our findings contribute to the advancement of predictive modeling techniques in bicycling volume estimation and offer insights for city planners and policymakers aiming to enhance bicycling infrastructure and promote sustainable transportation choices.

The remaining paper is structured as follows: \autoref{sec:litreview} presents a detailed review of the existing literature on bicycling volume estimation and the application of machine learning and deep learning models in traffic prediction with a focus on the challenges posed by data sparsity. 
In \autoref{sec:dataset}, we describe the datasets used in this study including the OpenStreetMap (OSM) network data and Strava Metro data, and emphasize on the relevance and integration of these datasets in our research. 
Section~\ref{sec:methodology} sets out the details of our methodology including data preparation process, the construction of the graph from the network data, and the design of the Graph Convolutional Network (GCN) architecture employed in this study. Additionally, it describes the training process and the systematic simulation of varying levels of data sparsity to assess the impact on the GCN model’s performance.
In \autoref{sec:results}, we present and discuss the results, including the identification of the optimal GCN configuration, finding the optimal hyperparameters for the machine learning models, and the analysis of how data sparsity effects the performance of GCN and machine learning models. 
Finally, \autoref{sec:conclusion} concludes the paper, summarizing key findings and suggesting direction of future research to improve link-level bicycling volume estimation modeling.

\section{Literature Review} \label{sec:litreview}

The literature on traffic volume prediction has significantly advanced with researchers focused on exploring innovative methodologies to enhance prediction accuracy. 
Among these methodologies, Graph Neural Networks (GNNs) have emerged as a powerful tool for modeling complex relationships within graph-structured data, such as road networks.
Road networks being inherently graph-like are naturally suited for GNN applications. Prominent approaches such as Diffusion Convolutional Recurrent Neural Networks (DCRNN) by \citep{li2017diffusion} and Spatio-Temporal Graph Convolutional Networks (ST-GCN) by \citep{yu2017spatio} have set benchmarks in motorised traffic volume prediction. 
DCRNN excels in capturing both spatial and temporal dependencies through its diffusion convolution operation, which enables it to model long-range spatial interactions and evolving traffic patterns more effectively than traditional methods. 
Similarly, ST-GCN integrates spatial and temporal dimensions by constructing dynamic graphs that connect nodes across time steps, thereby enhancing its ability to estimate traffic flow in complex urban environments. 
While DCRNN and ST-GCN have demonstrated robustness in motorized traffic volume estimation, their direct application to bicycling traffic presents unique challenges due to irregular patterns, infrastructure variability, and data sparsity. 
Both motorized and bicycling traffic systems face the challenge of incomplete or missing data, which limits the effectiveness of traditional methods. 
In the case of bicycling traffic, data sparsity is particularly problematic due to the less consistent use of infrastructure and diverse user behaviors, which complicates accurate link-level volume estimation. 

GNNs offer a promising solution to mitigate the effects of data sparsity with their ability to handle irregularly structured data.
Recent advancements such as the GraphSAGE architecture proposed by \citep{liu2020graphsage}, have demonstrated how GNNs can aggregate information from neighboring nodes to compensate for missing data. 
Additionally, the Dynamic Traffic Graph Neural Network (DTGNN) introduced by \citep{lei2022modeling} adapts to real-time traffic conditions by dynamically updating edge weights based on current traffic signals further addressing the data sparsity sparsity challenges. 
These studies highlight the GNNs potential in enhancing the robustness of motor traffic volume predictions, particularly under sparse data conditions.

Despite these advancements in motorized traffic volume prediction, the application of GNNs to bicycling volume estimation remains under-explored. 
Traditional bicycling volume estimation models, such as Ordinary Least Squares (OLS) \citep{hankey2012estimating}, log-linear regression \citep{strauss2013spatial, el2018daily}, stepwise linear regression \citep{lu2018adding, hochmair2019estimating}, generalized linear model (GLM) \citep{chen2017built}, poisson regression count \citep{fagnant2016direct, el2018daily}, Gaussian models \citep{ermagun2018bicycle} and Bayesian modeling \citep{strauss2013cyclist}, have been employed in earlier studies but often struggle to capture the non-linear and complex patterns inherent in bicycling data. 
While these models are interpretable, their limitations become apparent when applied to the dynamic and irregular nature of bicycling traffic. 
Altough the advent of machine learning and deep learning models \citep{miah2023estimation, das2020interpretable}, including Support Vector Machines (SVM), Random Forests, and Convolutional Neural Networks (CNNs) offer some advancements, they remain limited in their ability to effectively process the complex relationships within transport network data especially with data sparsity. 

In recent years, Graph Neural Networks (GNNs) have gained significant attention for their capability to model intricate relationships within graph-structured data, making them particularly well-suited for applications in road networks. 
While there has been some research in the bicycle domain, such as the work of \citep{lin2018predicting, lin2020predicting, yang2020using}, it has predominantly focused on modeling bike-sharing system volumes. 
Bike-sharing systems are public bicycle rental programs that allow users to rent bikes for short-term trips through a network of docking stations. 
These systems generate detailed data such as trip start and end points, rental duration, and user demographics which enables accurate prediction of bike-sharing counts. 
In contrast, link-level bicycling volume estimation involves predicting bicycle traffic for all bicyclists on all street segments across a city. 
This task is more complex due to factors like data sparsity, geographic spread and complex mobility patterns. 
These challenges highlight why traditional machine learning models while successful in bike-sharing contexts struggle with the complexities of link-level bicycling volume estimation. 
The sparsity, variability, and distribution of bicycle traffic data may benefit from advanced techniques like Graph Neural Networks (GNNs) which can capture spatial dependencies and address the challenges presented by irregular and incomplete data. To bridge this gap, our study proposes an efficient GCN architecture specifically designed for link-level bicycling volume modeling. 

\section{Datasets used} \label{sec:dataset}
Our research is focused on the geographical region of the City of Melbourne, a densely populated metropolitan area surrounding and including the central business district (CBD) of Melbourne, Australia. 
It covers an area of approximately 37.7 square kilometers and is characterized by a dense network of roads, pathways, and dedicated bicycle infrastructure. 
It includes a network of on-road bike lanes and off-road shared paths, comprising a total of 15,933 individual links that connect various parts of the city. 
The high density and diverse transportation infrastructure make Melbourne an ideal location for studying bicycling patterns and predicting link-level bicycling volumes. 

However, ground-truth observations of bicycling counts cover only a small fraction of these links, underscoring the substantial data sparsity within the network \citep{costa2020challenges, nelson2021crowdsourced}. To achieve a comprehensive understanding of the bicycling network within the City of Melbourne, we rely on infrastructure data from OpenStreetMap (OSM) and crowd-sourced bicycle volume data from Strava Metro in this study.

\subsection{OSM Network Data}
OSM \citep{OpenStreetMap,haklay2008openstreetmap} is an open-source data that serves as a foundational dataset for our study, providing a detailed and comprehensive representation of the transportation network within the study area. 
With our specific focus on bicycle-related infrastructure, we extract the road/path network data from OSM which includes the topology and geometry of the road network, as well as associated features such as slope, the presence and location of dedicated bicycle infrastructure such as bike lanes, shared road spaces. 
This data allows us to construct the graph representation of the network, where nodes represent road segments/links, and edges represent the intersection between them.
By integrating this network data with our bicycle count dataset, we adopt a graph-based modeling approach to achieve accurate bicycling volumes modeling and a more informed analysis of bicycle network  across the city.

\subsection{Strava Metro Data}

This paper utilizes aggregated and de-identified data from Strava Metro \citep{StravaMetro}, a widely used fitness tracking app that has collected a vast repository of cycling activity data. 
Strava Metro captures detailed cycling information, including route data, ride frequency, and the spatial and temporal distribution of cyclist movements.

At the individual level, the raw data includes specifics such as trip start and end locations, timestamps, and routes taken. 
However, Strava Metro also provides aggregate-level data that reflects cyclist volumes on each link in the network, segmented by factors such as time of day and day of the week. 
This rich dataset with its temporal and spatial granularity, allows for more accurate modeling of link-level bicycling volumes. 
It has been employed as a key explanatory variable in multiple bicycling-related studies \citep{selala2016potential, lee2020emerging, ferster2021mapping, nelson2023bicycle}.

Unlike traditional limited count data, Strava Metro offers bicycling counts across all 15,933 links in the network. 
This extensive coverage enables us to simulate various levels of data sparsity by systematically reducing the available count data and examining its impact on the performance of our Graph Convolutional Network (GCN) model. 
In other words, we train the GCN on a fully populated Strava dataset, then progressively remove portions of the data to replicate real-world data scenario, evaluating how the model's performance degrades under more extreme sparsity conditions.

Despite its extensive coverage, it is important to recognize the limitations of Strava Metro data. 
It tends to over-represent specific cyclist groups, such as recreational riders and fitness enthusiasts, which can introduce bias by emphasizing well-trafficked routes and underreporting quieter or less-traveled areas. 
This bias may skew the overall representation of cycling behavior \citep{nelson2021generalized, venter2023bias}. 
Nevertheless, the comprehensive nature of Strava Metro data serves as a valuable testbed for understanding how data sparsity affects link-level bicycling volume estimation under realistic conditions of incomplete observation or limited bicycle count data.

\section{Methodology} \label{sec:methodology}
\subsection{Data preparation and processing}

\subsubsection{Extracting the Bicycle Network from OSM}

As the first step, we extracted the road/path network of the City of Melbourne from OpenStreetMap (OSM) using the OSMnx \citep{boeing2017osmnx} library in Python. 
The OSM dataset provides unique OSM IDs for each feature, distinguishing nodes, ways, and relations, along with detailed attributes for each road/path, such as surface type, and speed limit. 
To ensure comprehensive coverage of all roads/paths where bicyclists are permitted to ride, we extracted relevant OSM data using OSMnx by employing a robust combination of tags and corresponding values.
We then used a bike infrastructure classification system \citep{Sustainable_Mobility_and_Safety_Research_Group_Bicycling_infrastructure_classification_2023} and bike level of traffic stress (LTS) model \citep{bike_lts} bespoke to Greater Melbourne to appropriately classify the types of infrastructure and bicycle level of traffic stress for each street segment in the network, respectively.
\subsubsection{Calculating Strava AADB}
To standardize the data and account for both temporal and spatial variations, we calculated Annual Average Daily Bicycle (AADB) count for each OSM ID by aggregating and averaging the daily bicycle counts associated with that road/path segments using the equation:
\begin{equation}\label{aadb_eq}
    AADB_i=\left\lceil\frac{\sum_{j=1}^{n_i}{bicycle\_trip\_count}_{i j}}{n_{i}}\right\rceil
\end{equation}
where $AADB_{i}$, $bicycle\_trip\_count_{ij}$ and $n_{i}$ represent the annual average daily bicycle count, count for the $j^{th}$ observation, total number of observations for the $i^{th}$ OSM ID respectively and $\lceil . \rceil$ denotes the ceiling function, rounding up to the nearest integer.
Strava AADB offers a consistent measure for comparing bicycling volumes across diverse locations, smoothing out fluctuations caused by seasonal effects or short-term anomalies. Although it derives from Strava data, which can be biased toward specific cyclist groups (e.g., recreational or fitness-oriented riders), this measure captures a broad distribution of cycling activity across 15,933 segments in the City of Melbourne.

\subsubsection{Integrating OSM and Strava Data}
To form a comprehensive dataset for link-level bicycling volume modeling, we integrate the OSM-based road/path network with Strava AADB counts. Each segment is linked via a unique OSM identifier, ensuring consistency in how infrastructure attributes and cycling volumes are matched. This combined dataset serves as the foundation for training and evaluating the Graph Convolutional Network (GCN) approach used in our experiments.

While Strava provides counts for the entire network, real-world conditions often involve sparse or incomplete bicycle count data. To simulate these conditions, we progressively reduce the available Strava count data, creating varying levels of sparsity. This strategy allows us to analyze how model performance changes as data becomes more limited, reflecting challenges faced by urban planners in cities with few permanent counters or limited resources.

Before configuring the input for the Graph Convolutional Network (GCN) architecture, we conduct several data enrichment steps to ensure the dataset is well-prepared for modeling bicycling volumes. 
We supplement missing or inaccurately recorded attributes in OSM, such as slope and speed limits respectively, by spatially merging the OSM network with additional datasets which are as follows - \citep{slope_data} (slope data) and \citep{DTP_speed_data} (speed limit data), ensuring a more comprehensive representation of key features relevant to bicycling volumes on road segments.

\subsubsection{Data Enrichment and Cleaning}
Additionally, missing data in categorical variables such as road/path surface types and infrastructure classifications are imputed using the mode value, ensuring that the most common category filled any gaps while continuous variables such as segment length and speed limit are imputed using the mean, avoiding distortions caused by outliers. 
Given the diverse nature of the data, Min-Max scaling is applied to continuous predictors to normalize their values and ensure they are on a comparable scale. 
Categorical variables, which include classifications of road types and bicycle facilities, are converted into numerical format through one-hot encoding, enabling the GCN to process these features effectively. 

\subsubsection{Addressing Skewness in Strava AADB}

\begin{figure}[H]
    \centering
    \includegraphics[width=1\linewidth]{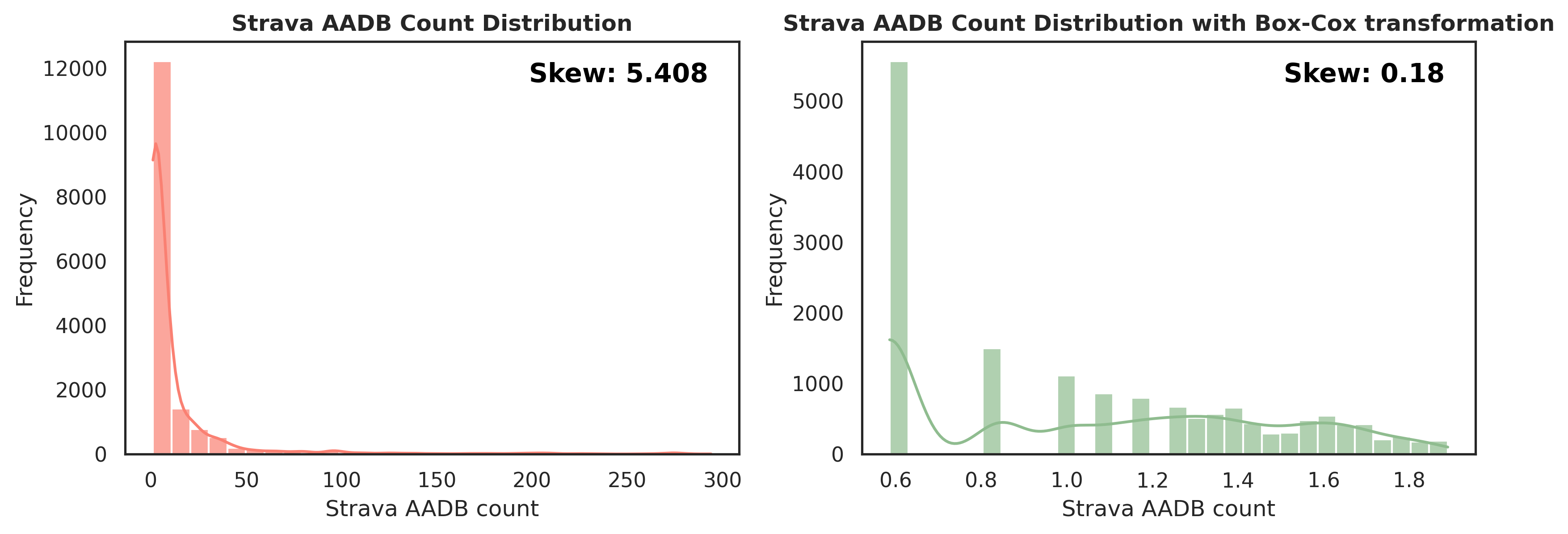}
    \caption{\small Comparison of Strava AADB Count Distribution Before and After Box-Cox Transformation}
    \label{fig:transformation}
\end{figure}

As shown in \autoref{fig:transformation}, the Strava AADB distribution exhibits a right (positive) skew, with most road segments having relatively low bicycle volumes while a small portion have much higher counts. 
Such skewness can lead to biased model predictions and limit the model’s generalization. 
To mitigate this issue, we test multiple transformations, including Log, Box-Cox, Square Root, Quantile and Yeo-Johnson.

\autoref{tab:Skewness} shows that Box-Cox and Yeo-Johnson transformations yield the most substantial reduction in skewness. 
Given its consistency, we select the Box-Cox transformation for our target variable in the GCN modeling task.

\begin{table}[H]
\centering
\caption{\small Skewness comparison for different data transformations applied on Strava AADB Count}
\label{tab:Skewness}
\renewcommand{\arraystretch}{1.5}
\begin{tabular}{l|cccccc}
                  & \multicolumn{6}{c}{\textbf{Transformation}}                                                                     \\ \hline
                  & \textbf{No} & \textbf{Square Root} & \textbf{Log} & \textbf{Quantile} & \textbf{Yeo-Johnson} & \textbf{Box-Cox} \\
\textbf{Skewness} & 5.408       & 2.674                & 0.74         & -0.64             & 0.26                 & 0.18            
\end{tabular}
\end{table} 

\noindent
With these data preparation and preprocessing steps completed, the next step is to transform the data into graph structures that represent the spatial and infrastructure relationships within the bicycle network. 
The following section will delve into the graph preparation process and its critical role in enhancing the GCN model's ability to capture complex patterns in bicycling activity.

\begin{figure}[H]
    \centering
    \includegraphics[width=1\linewidth]{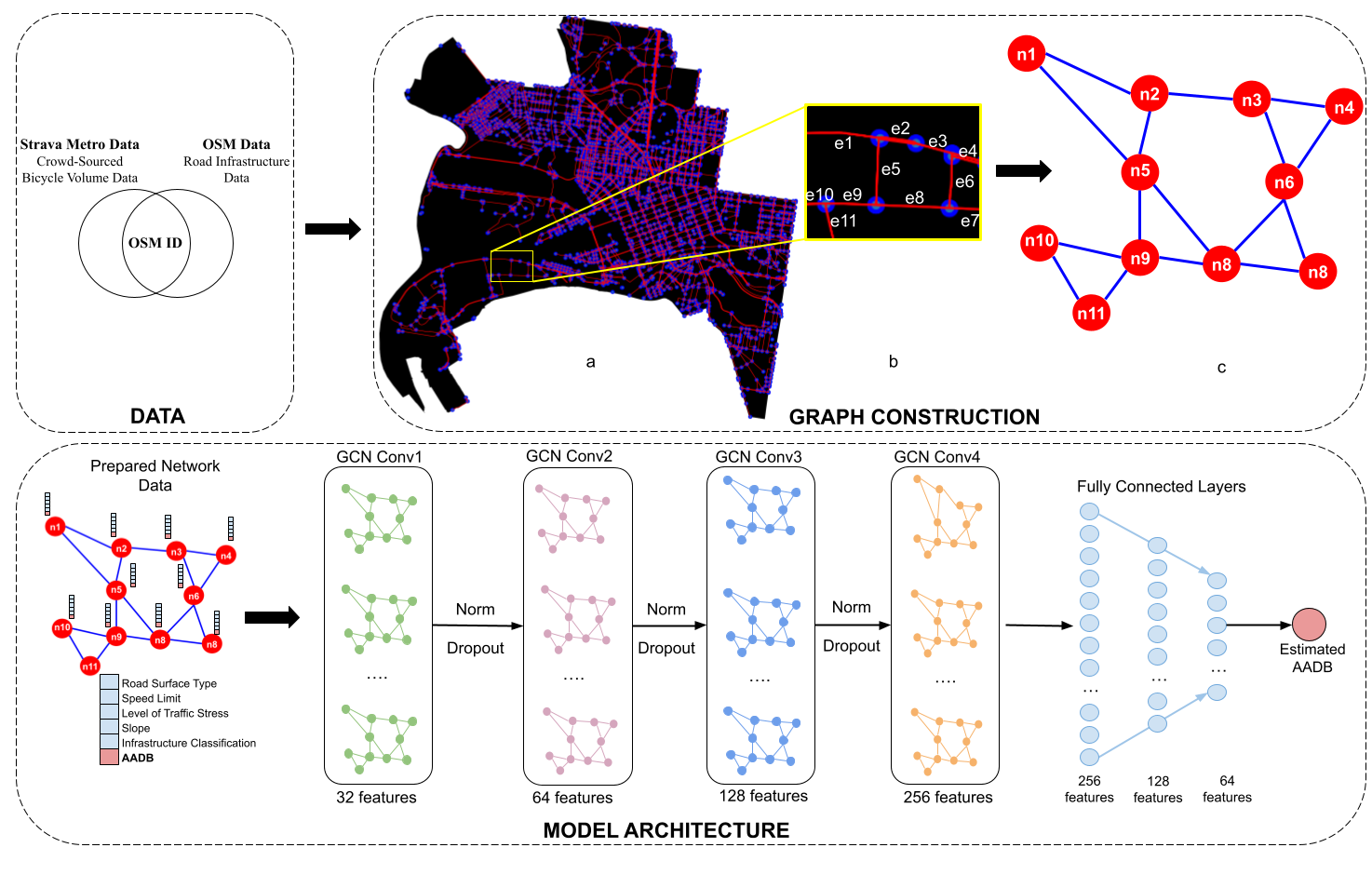}
    \caption {Overview of methodology for link-level bicycling volume prediction using using a Graph Convolutional Network (GCN) approach. Data section shows integration of Strava Metro Data (crowd-sourced bicycle volume data) and OpenStreetMap (OSM) Data (road infrastructure data), linked via unique OSM IDs. The graph construction phase includes: (a) mapping the bicycle network within the City of Melbourne, (b) demonstrating the process of graph inversion on a small area, where road segments are converted into nodes and intersections into edges, and (c) forming the final node-centric graph representation. The model architecture section is representing the GCN Configuration G (discussed in \ref{sec:gcn_config} and \ref{sec:gcn_config_result})}
    \label{fig:methodology_diagram}
\end{figure}

\subsection{Graph Creation for GCN}\label{sec:graph_creation}
This section details the construction of a graph representation of the prepared bicycling volume dataset, enabling the evaluation of data sparsity effects on link-level bicycling volume prediction using a Graph Convolutional Network (GCN). 
The graph creation process integrates data from OpenStreetMap (OSM) and Strava Metro, ensuring a spatially coherent and data-rich representation of Melbourne’s bicycling network.

\subsubsection{Inverting the Traditional OSM Representation}
A critical consideration in constructing this graph is the inversion of the traditional OSM representation, which represents intersections as nodes and road segments as edges. Instead, in our graph, each road/path segment is represented as a node, and intersections are modeled as edges connecting these nodes (as shown in \autoref{fig:methodology_diagram}). This node-centric approach allows for the direct assignment of bicycling volume estimates to individual road segments. Moreover, this inversion aligns with graph-based framework, enabling the GCN to effectively learn from both segment attributes and their spatial relationships.

\subsubsection{Graph Construction Process}
The graph construction involves following key steps:
\begin{itemize}
    \item \textbf{Defining Nodes and Features:} Each road or path segment, uniquely identified by its OSM ID, is represented as a node. 
    Node features include attributes such as road type, surface quality, speed limit, level of traffic stress, slope, and bicycle infrastructure classification. 
    Additionally, the Strava-derived Annual Average Daily Bicycle (AADB) count is assigned to each node as the target variable for prediction. 
    These features provide rich contextual information, enabling the GCN to leverage both intrinsic road characteristics and network-wide relationships.
    
    \item \textbf{Establishing Edges and Relationships:} Edges represent intersections or physical connections between road segments. 
    If two segments share a common intersection in OSM, an edge is created between their corresponding nodes. 
    Edge features capture spatial relationships such as shared infrastructure types, allowing the GCN to model dependencies within the network effectively.
    
    \item \textbf{Integration with PyTorch Geometric:} The final graph is formatted for compatibility with PyTorch Geometric, comprising node feature tensors, edge indices, and edge attribute tensors. 
    This structure enables the GCN to utilize both node attributes and edge relationships to predict bicycling volumes based on the intrinsic properties of the road segments and their network context.
\end{itemize}

\noindent
The node-centric graph structure is therefore pivotal for several reasons - 1. It preserves spatial relationships between road segments, allowing the model to learn how bicycle volumes on connected segments influence each other, 2. it facilitates the use of libraries like PyTorch Geometric, optimized for node regression tasks. It allows for controlled masking of nodes to simulate varying levels of data sparsity, providing insights into model robustness under realistic data conditions. This graph structure serves as the foundation for the GCN, enabling it to aggregate information from connected nodes and learn complex patterns in bicycling activity. The subsequent section details the training process and how the GCN leverages this graph structure to learn and generalize across the network.

\subsection{Graph Convolutional Network (GCN)}
The foundational concept of Graph Convolutional Networks (GCNs) shares similarities with ordinary feed-forward artificial neural networks \citep{hassoun1995fundamentals}, yet GCNs are distinct due to their ability to model relationships within graph-structured data, such as bicycling networks. 
Unlike traditional Multilayer Perceptrons (MLPs) where each neuron is fully connected to all neurons in the preceding layer, GCNs exploit the spatially local correlations inherent in graph data. 
In GCNs, each neuron connects only to a small region of the preceding layer, known as the receptive field, which represents the neighboring nodes in the graph.
In the context of our study, each node corresponds to a road/path segment, and edges represent the physical connections between these segments in the bicycling network. 
The key idea is to learn a representation of each segment based on both its own attributes and the attributes of its neighboring segments. This learned representation is then utilized to estimate bicycling volumes on these segments.
GCNs employ a series of graph convolution layers where, at each layer, node features are aggregated and transformed using the features of their neighboring nodes. 
This iterative process allows the model to progressively capture more complex relationships within the bicycling network, leading to improved feature extraction and predictive performance. 
The integration of GCN in this study enables the effective modeling of both local and global spatial dependencies, crucial for accurate link-level bicycling volume prediction.
The following sections detail the specific components of the GCN architecture used in this study, including the input layer, graph convolutional layers, activation functions, batch normalization, regularization techniques, and fully connected layers, as well as the various GCN configurations explored during the research. 

\subsubsection{Input Layer}
The input layer of the Graph Convolutional Network (GCN) is responsible for processing the raw features associated with each node in the graph, representing road or path segments, and preparing them for the subsequent convolutional operations. Let \( \mathbf{X} \) be the input feature matrix where $\mathbf{X} \in R^{N \times F}$. Here, \( N \) is the number of nodes (i.e., road segments) in the graph, and \( F \) is the number of features associated with each node. 
The input layer performs an initial transformation on the node features before they are passed through the graph convolutional layers. This can be expressed as:
\begin{equation}
    \mathbf{H}^{(0)} = \mathbf{X}
\end{equation}

where \( \mathbf{H}^{(0)} \) represents the initial node feature matrix, which is identical to the input feature matrix \( \mathbf{X} \) at this stage. The features in \( \mathbf{H}^{(0)} \) are directly passed to the next layer in the network without any transformation.
Each feature contributes to the overall representation of the nodes, enabling the GCN to capture both the intrinsic properties of the road segments and their relational context within the bicycling network. 
The input layer ensures that the features are structured in a way that leverages the connectivity of the graph, allowing the GCN to effectively model the spatial dependencies inherent in the data.

\subsubsection{Graph Convolutional (GCNConv) Layer}

The Graph Convolutional (GCNConv) layer is the core component of the GCN architecture, designed to capture the spatial relationships between nodes in a graph. 
The GCNConv layer updates the representation of each node by aggregating features from its neighbors and itself, thereby enabling the model to learn meaningful patterns within the graph structure. 
Mathematically, the operation performed by a single graph convolutional layer can be expressed as:
\begin{equation}
\mathbf{H}^{(l+1)} = \sigma \left( \tilde{\mathbf{D}}^{-\frac{1}{2}} \tilde{\mathbf{A}} \tilde{\mathbf{D}}^{-\frac{1}{2}} \mathbf{H}^{(l)} \mathbf{W}^{(l)} \right)
\end{equation}
where, 
$\mathbf{}{H}^{(l)} \in \mathbb{R}^{N \times F_l}$ represents the node features at layer $l$, $N$ is the number of nodes and $F_l$ is the number of features in layer $l$ and $\mathbf{H}^{(l+1)} \in \mathbb{R}^{N \times F_{l+1}}$ is the node features at layer $l+1$, with $F_{l+1}$ features. 
$\tilde{\mathbf{A}} = \mathbf{A} + \mathbf{I}$ is the augmented adjacency matrix, where $\mathbf{A} \in \mathbb{R}^{N \times N}$ is the original adjacency matrix and $\mathbf{I}$ is the identity matrix added to include self-loops. $\tilde{\mathbf{D}}$ is the diagonal degree matrix of $\tilde{\mathbf{A}}$, where each element $\tilde{d}_{ii}$ represents the degree of node $i$.
$\mathbf{W}^{(l)} \in \mathbb{R}^{F_l \times F_{l+1}}$ is a learnable weight matrix at layer $l$, which is used to transform the features from layer $l$ to layer $l+1$.
$\sigma(\cdot)$ denotes the activation function applied element-wise after the convolution operation.

The operation $\tilde{\mathbf{D}}^{-\frac{1}{2}} \tilde{\mathbf{A}} \tilde{\mathbf{D}}^{-\frac{1}{2}}$ performs a normalization of the adjacency matrix, which ensures that the features are scaled appropriately and prevents issues such as exploding or vanishing gradients during training. 
The learnable weight matrix $\mathbf{W}^{(l)}$ is crucial for adapting the feature space to the specific task of predicting link-level bicycling volumes. 
The activation function introduces non-linearity into the model, allowing the GCN to learn complex patterns and relationships within the data. After each GCNConv layer, an activation function $\sigma(\cdot)$ is applied element-wise to the output. 
By stacking multiple GCNConv layers, the GCN can learn increasingly abstract representations of the graph data which enables it to capture both local and global structures within the bicycling network.

\subsubsection{Activation Layer, Batch Normalization, and Regularization}
In the Graph Convolutional Network (GCN) architecture, the combination of activation functions, batch normalization, and regularization techniques plays a pivotal role in enhancing the model's performance, stability, and generalization ability.

We have used the Rectified Linear Unit (ReLU) activation function, which is defined as:
 \begin{equation}
     \sigma(x) = \text{ReLU}(x) = \max(0, x) 
 \end{equation}
The ReLU activation function is chosen due to its simplicity and effectiveness in addressing the vanishing gradient problem which otherwise can hinder the training of our GCN.

Batch normalization (BatchNorm) is employed after the activation layer to stabilize and accelerate the training process. BatchNorm normalizes the output of the previous layer by adjusting and scaling the activations, ensuring that they have a consistent distribution throughout training. Mathematically, BatchNorm for a given mini-batch is expressed as:
\begin{equation}
    \hat{x}_i = \frac{x_i - \mu_B}{\sqrt{\sigma_B^2 + \epsilon}}, \quad y_i = \gamma \hat{x}_i + \beta
\end{equation}

where,
$x_i$ is the input feature for the $i^{th}$ node in the mini-batch.
$\mu_B$ and $\sigma_B^2$ are the mean and variance of the mini-batch, respectively.
$\epsilon$ is a small constant added for numerical stability.
$\gamma$ and $\beta$ are learnable parameters that scale and shift the normalized output.
BatchNorm helps in mitigating issues such as internal covariate shift where the distribution of layer inputs changes during training and allows for higher learning rates to reduce the number of training epochs required to reach convergence. 

Regularization techniques are critical in preventing overfitting, especially when the model is trained on sparse data, as in the case of bicycling volume prediction. 
In this GCN architecture, we incorporate dropout regularization after the activation and batch normalization layers.
Dropout randomly sets a fraction of the activations in the layer to zero during each training iteration, which forces the network to learn redundant representations and prevents over-reliance on any particular set of neurons. The dropout operation is mathematically represented as:
\begin{equation}
    y_i^{\text{drop}} = \begin{cases} 0 & \text{with probability } p, \\ y_i & \text{with probability } 1-p, \end{cases}
\end{equation}
where $y_i$ is the output after BatchNorm and $p$ is the dropout probability. Using grid search method, we have set the optimal value of $p$ to 0.4 in our GCN architecture. 
The combination of these ReLU activation, BatchNorm, and dropout techniques ensures that the GCN model can learn robust features from the graph-structured data while handling the risk of overfitting and improving the generalization to unseen data.

\subsubsection{Fully Connected Layer}
The Fully Connected (FC) layer, also known as a dense layer, serves as the final component of the Graph Convolutional Network (GCN) architecture. 
It plays a crucial role in integrating the high-level features extracted from the graph convolutional layers and producing the final predictions for link-level bicycle volumes.
In the FC layer, each node's representation, which has been progressively refined through the graph convolutional and other intermediate layers, is mapped to the output space. This is achieved by connecting each node's features to every unit in the output layer, ensuring that all features contribute to the final prediction.
Mathematically, if \( \mathbf{h}_i^{(L)} \) represents the hidden representation of node \( i \) from the last graph convolutional layer (where \( L \) is the number of layers), the fully connected layer transforms this representation into the final output using a weight matrix \( \mathbf{W}_{\text{FC}} \) and a bias term \( \mathbf{b}_{\text{FC}} \). The operation can be expressed as:
\begin{equation}
    \mathbf{z}_i = \mathbf{W}_{\text{FC}} \mathbf{h}_i^{(L)} + \mathbf{b}_{\text{FC}}
\end{equation}
where, \( \mathbf{z}_i \) is the output vector for node \( i \), which can correspond to the predicted Annual Average Daily Bicycle (AADB) count for the given road/path segment. 
\( \mathbf{W}_{\text{FC}} \) is the weight matrix of the fully connected layer, determining how each feature contributes to the output.
\( \mathbf{b}_{\text{FC}} \) is the bias vector that shifts the output.

The output \( \mathbf{z}_i \), predicted \( \hat{AADB} \) value is then direclty compared with the actual AADB counts during the training process to compute the loss.

\subsubsection{GCN Configuration} \label{sec:gcn_config}
To identify the optimal Graph Convolutional Network (GCN) architecture for estimating link-level bicycle volumes, we explore a diverse range of configurations labeled A through J. 
These configurations differs in depth, layer composition, and architectural enhancements as mentioned in Table \ref{tab:confguration}. 
Each configuration is meticulously designed by varying several architectural components - the number of Graph Convolutional (GCNConv) layers, the inclusion of Batch Normalization, Dropout layers, and Fully Connected (FC) layers.
The configurations, representing distinct combinations of key architectural choices are designed to assess the impact of different choices on the model’s performance, particularly in terms of capturing the complex spatial dependencies inherent in bicycling networks.

The configurations range from simpler models (A-D) with fewer feature channels (32, 64) in the GCNConv layers to more complex architectures (E-J) with deeper networks and increased feature channels (128, 256, and 512). 
Configurations A, B, and C represent the baseline architectures where the GCNConv layers consist of 32 and 64 feature channels. 
These baseline models provide a benchmark for understanding how minimal architectures perform in comparison to more advanced configurations.

\begin{table}[H]
\centering
\caption{\small Overview of the different GCN configurations (A-J) highlighting key architectural variations. [Bold text indicates where the change started from in the layer configuration compared to the previous configuration. Everything before the bold text is common in both the configurations.]}
\label{tab:confguration}
\renewcommand{\arraystretch}{1.4}  
\begin{tabular}{>{\raggedright\arraybackslash}p{0.2\linewidth}|%
>{\centering\arraybackslash}p{0.04\linewidth}>{\centering\arraybackslash}p{0.04\linewidth}>%
{\centering\arraybackslash}p{0.04\linewidth}>{\centering\arraybackslash}p{0.04\linewidth}>%
{\centering\arraybackslash}p{0.04\linewidth}>{\centering\arraybackslash}p{0.04\linewidth}>%
{\centering\arraybackslash}p{0.04\linewidth}>{\centering\arraybackslash}p{0.04\linewidth}>%
{\centering\arraybackslash}p{0.04\linewidth}>{\centering\arraybackslash}p{0.04\linewidth}}
\textbf{Layer} & \textbf{A} & \textbf{B} & \textbf{C} & \textbf{D} & \textbf{E} & \textbf{F} & \textbf{G} & \textbf{H} & \textbf{I} & \textbf{J} \\ \hline
GCNConv               & 32 & 32           & 32           & 32           & 32           & 32           & 32           & 32           & \textbf{64} & 64           \\
GCNConv               & 64 & 64           & 64           & 64           & 64           & 64           & 64           & 64           & 128         & 128          \\
BatchNorm             & No & \textbf{Yes} & Yes          & Yes          & Yes          & Yes          & Yes          & Yes          & Yes         & Yes          \\
Dropout               & No & Yes          & Yes          & Yes          & Yes          & Yes          & Yes          & Yes          & Yes         & Yes          \\
GCNConv               & No & No           & No           & \textbf{128} & 128          & \textbf{No}  & \textbf{128} & 128          & 256         & 256          \\
BatchNorm             & No & No           & No           & No           & \textbf{Yes} & No           & Yes          & Yes          & Yes         & Yes          \\
Dropout               & No & No           & No           & No           & Yes          & No           & Yes          & Yes          & Yes         & Yes          \\
GCNConv               & No & No           & No           & No           & No           & 256          & 256          & 256          & No          & \textbf{512} \\
BatchNorm             & No & No           & No           & No           & No           & Yes          & No           & \textbf{Yes} & No          & Yes          \\
Dropout               & No & No           & No           & No           & No           & Yes          & No           & Yes          & No          & Yes          \\
Fully Connected Layer & No & No           & \textbf{64}  & 128          & 128          & 256          & 256          & 256          & 256         & 512          \\
Dropout               & No & No           & Yes          & Yes          & Yes          & Yes          & Yes          & Yes          & Yes         & Yes          \\
Fully Connected Layer & 64 & 64           & 64           & 64           & 64           & 128          & 128          & 128          & 128         & 128          \\
Dropout               & No & No           & No           & No           & No           & Yes          & Yes          & Yes          & Yes         & Yes          \\
Fully Connected Layer & No & No           & No           & No           & No           & 64           & 64           & 64           & 64          & 64          
\end{tabular}
\end{table}

In configurations E through J, we progressively increased the depth and feature channels to capture more nuanced spatial relationships within the data. 
Configuration E adds an additional GCNConv layer with 128 feature channels while configurations G, H, I, and J go further by increasing the number of feature channels to 256 and 512 in later layers. 
These configurations aim to enhance the model’s feature extraction capability, especially in the presence of complex spatial interactions within the bicycling network.

Batch Normalization and Dropout is strategically applied across different configurations to mitigate overfitting and stabilize training. 
Batch Normalization is included in configurations with additional layers (C, E, F, G, H, I, and J) to standardize the input distributions, ensuring smoother gradients during training.
Dropout layers are included to randomly deactivate a fraction of neurons, particularly in configurations D through J, as the network complexity increased. 
The dropout mechanism helps to prevent overfitting by forcing the model to generalize better during the training process. 
Finally, Fully Connected (FC) layers are included in configurations with varying numbers of units (64 to 512), which allow us to examine their role in further refining the predictions from the convolutional layers. 
The FC layers aggregate information from the GCNConv layers and produce the final output for link-level volume predictions.

By designing a broad spectrum of GCN configurations, we ensure sufficient architectural variation to thoroughly evaluate the performance of the model under different conditions. 
These experiments are critical for understanding how each element affects the GCN’s ability to model bicycling volumes. By systematically testing these configurations, we aim to find the optimal balance between model complexity and performance and ensure that the model is robust enough to handle the challenges posed by data sparsity for the accurate bicycling volume estimation. 

\subsection{Training Process}
\begin{figure}[H]
    \centering
    \includegraphics[width=0.75\linewidth]{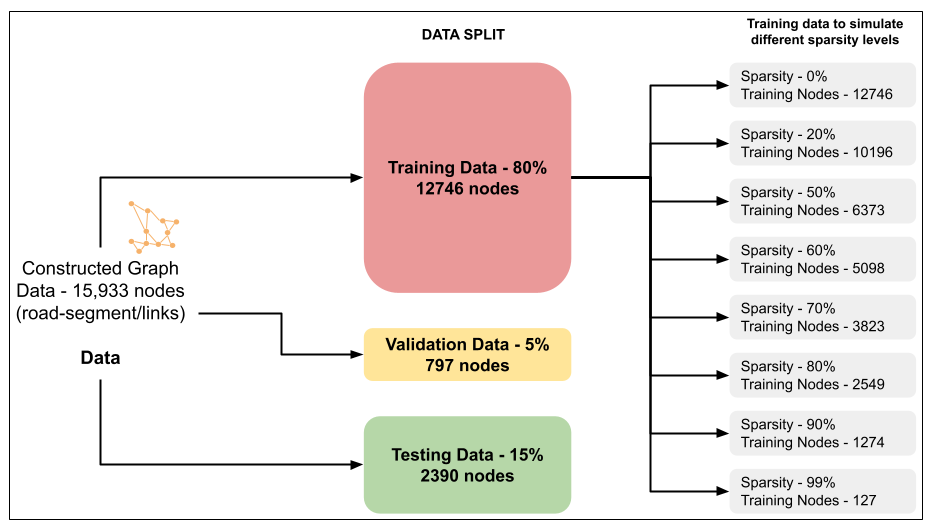}
    \caption{Data split and sparsity simulation for the training process. The constructed graph consists of 15,933 nodes (road segments/links), divided into training (80\% - 12,746 nodes), validation (5\% - 797 nodes), and testing (15\% - 2,390 nodes) sets. To simulate varying sparsity levels, the training data is progressively reduced from 0\% sparsity (12,746 nodes) to 99\% sparsity (127 nodes), enabling a systematic evaluation of model performance under different levels of data sparsity.}
    \label{fig:data_split}
\end{figure}

The training process of our Graph Convolutional Network (GCN) is designed to optimize the model for estimating link-level bicycle volumes with a focus on handling the inherent challanges of data sparsity and maximizing the use of spatial relationships within the network. 
Initially, the GCN architecture is defined by specifying the number of input features, hidden layer configurations and output dimensions that align with the characteristics of the graph-structured data derived from OpenStreetMap (OSM) and Strava Metro data. 

To train the GCN, we employ a Mean Squared Error (MSE) loss function which minimizes the difference between the predicted Annual Average Daily Bicycle (AADB) counts and the ground truth values. 
MSE is calculated using formula shown in \autoref{metrics} where, \({y_i}\) represents the true AADB count and \({\hat{y}_i}\) is the predicted count.
The MSE is chosen for its effectiveness in penalizing larger errors, thereby ensuring that significant deviations in predictions are minimized. 
Optimization of the model is performed using the Adam optimizer, which is well-suited for training deep learning models due to its adaptive learning rate mechanism that accelerates convergence while preventing overshooting the minima.
The learning rate is initialized at $10^{-3}$, and we incorporate weight decay (L2 regularization) to prevent overfitting by penalizing the large weights. 

Data normalization is a critical component of our preprocessing pipeline. Continuous features are normalized using Min-Max scaling to ensure they are on a comparable scale, which is crucial for stable training of neural networks. 
The target variable, AADB, undergoes Box-Cox transformation to address significant skewness and improve model generalizability. 
The categorical variables are transformed via one-hot encoding, allowing the GCN to effectively process both numerical and categorical data.

We divide the dataset into training, validation, and test sets with a split ratio of 80:5:15 to comprehensively evaluate model performance as shown in \autoref{fig:data_split}. 
The training dataset is used to learn model parameters, while the validation set is employed to tune hyperparameters and prevent overfitting. 
We have set 2500 epochs for model training with early stopping, which halts the training if the validation loss does not improve over a set number of epochs, preventing the model from overfitting and saving computational resources. 
This technique ensures that the model's capacity does not exceed the complexity of the data, especially in sparse settings.

Regularization techniques are extensively utilized to further mitigate overfitting. 
Dropout layers with probability 0.4 are placed after each GCN layer, randomly deactivating a portion of feature channels (neurons) during training to prevent the model from becoming too reliant on specific connections through the neural network.
Batch Normalization is incorporated to stabilize and accelerate the training process by normalizing activations and reducing internal covariate shifts which helps maintain consistent learning dynamics across epochs.

\begin{equation}\label{metrics}
\text{RMSE} = \sqrt{\frac{1}{N} \sum_{i=1}^{N} (y_i - \hat{y}_i)^2}, \quad \text{MSE} = \frac{1}{N} \sum_{i=1}^{N} (y_i - \hat{y}_i)^2, \quad \text{MAPE} = \frac{1}{N} \sum_{i=1}^{N} \left|\frac{y_i - \hat{y}_i}{y_i}\right| \times 100\%
\end{equation}

Throughout the training, we continuously monitor both training and validation losses to assess convergence and stability. 
We evaluate the model performance using several key metrics, including Root Mean Squared Error (RMSE), Mean Absolute Error (MAE), and Mean Absolute Percentage Error (MAPE). 
These metrics are calculated after every 50 epochs and at the end of training to assess both training and validation performance. 
The goal is to achieve a well-generalized model that performs consistently across both sets, with minimal performance degradation as data sparsity increases.

This comprehensive training process ensures that our GCN model is optimized not only for predictive accuracy but also for stability and generalization. It ensures that the model is well-suited for handling the challenges of sparse graph-structured data in the context of link-level bicycling volume prediction.

\subsection{Sparsity Simulation}
To systematically assess the impact of data sparsity on the performance of our GCN model, we introduced varying levels of sparsity into the dataset as shown in \autoref{fig:sparsity_simulation}. 
Sparsity is simulated by selectively masking a percentage of nodes in the training data i.e. intentionally removing a certain proportion of the labelled nodes with Strava AADB counts from model training. This is performed to mimic the real-world scenario where data collection might be incomplete and sparse. 

The removal process is performed randomly to ensure that the sparsity introduced is unbiased and reflective of potential missing data scenarios in bicycling networks. We have applied different sparsity levels - 20\%, 50\%, 60\%, 70\%, 80\%, 90\% and 99\%, to create datasets with progressively fewer data points, ultimately reaching a condition that mirrors the ground truth scenario in the City of Melbourne.
For each sparsity level, we recalculate the connectivity and structure of the graph to ensure that the remaining data still forms a coherent network. 
These sparse dataset is then used to train and evaluate the GCN model, allowing us to observe how the model's predictive performance is affected with increasing sparsity.      

By introducing controlled sparsity, we can systematically analyze its effects and compare the resilience of the GCN model against traditional ML models under similar conditions. This step is critical in understanding the robustness of the GCN architecture under conditions of incomplete data, which is a common challenge in urban planning and infrastructure development. 

\begin{figure}[H]
    \centering
    \begin{minipage}{.35\textwidth}
        \centering
        \includegraphics[width=\textwidth]{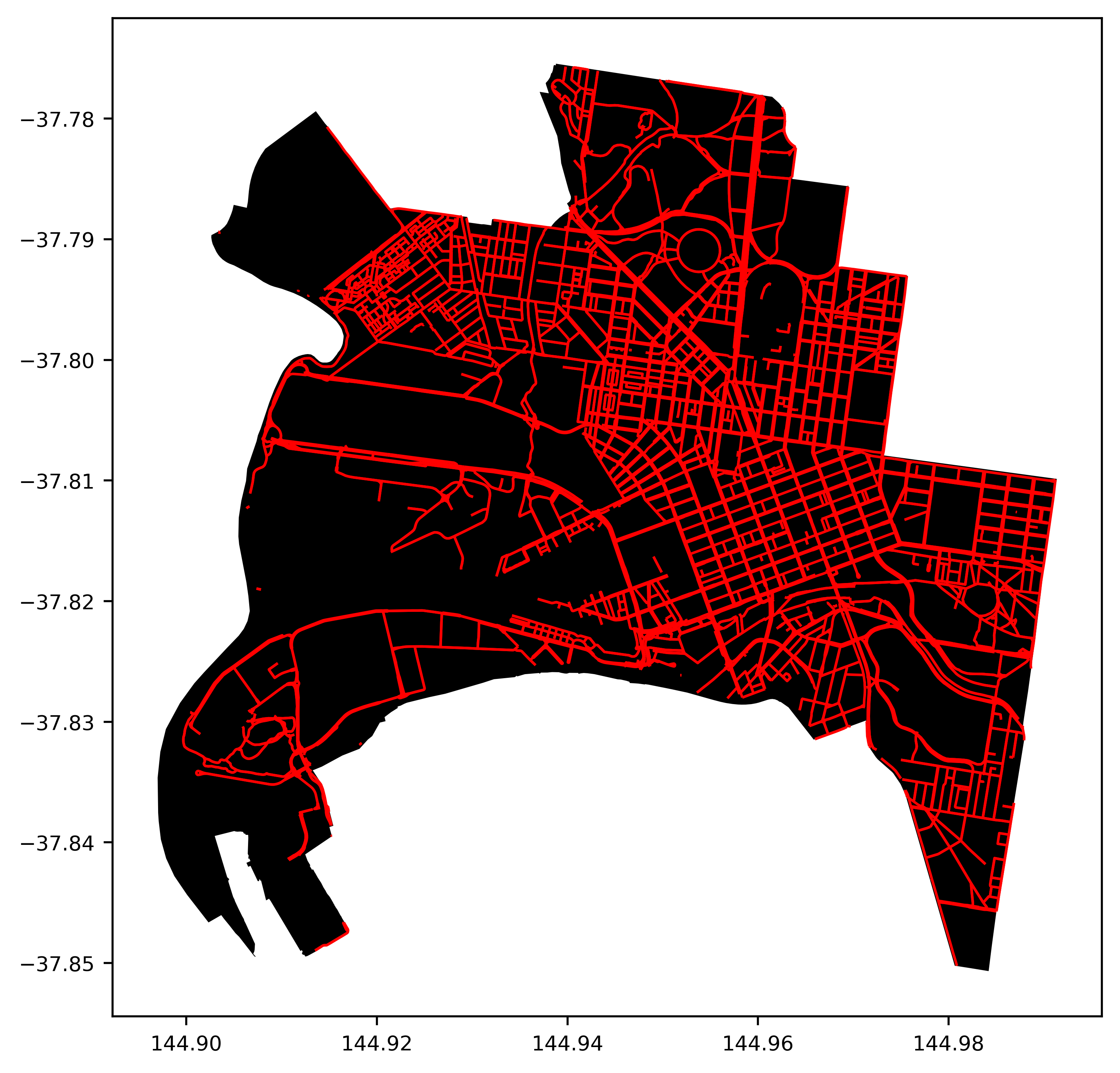}
        \subcaption{Bicycle Network in City of Melbourne}
    \end{minipage}
\end{figure}    

\begin{figure}[H]\ContinuedFloat
    \centering    
    \begin{minipage}{0.31\textwidth}
        \centering
        \includegraphics[width=\textwidth]{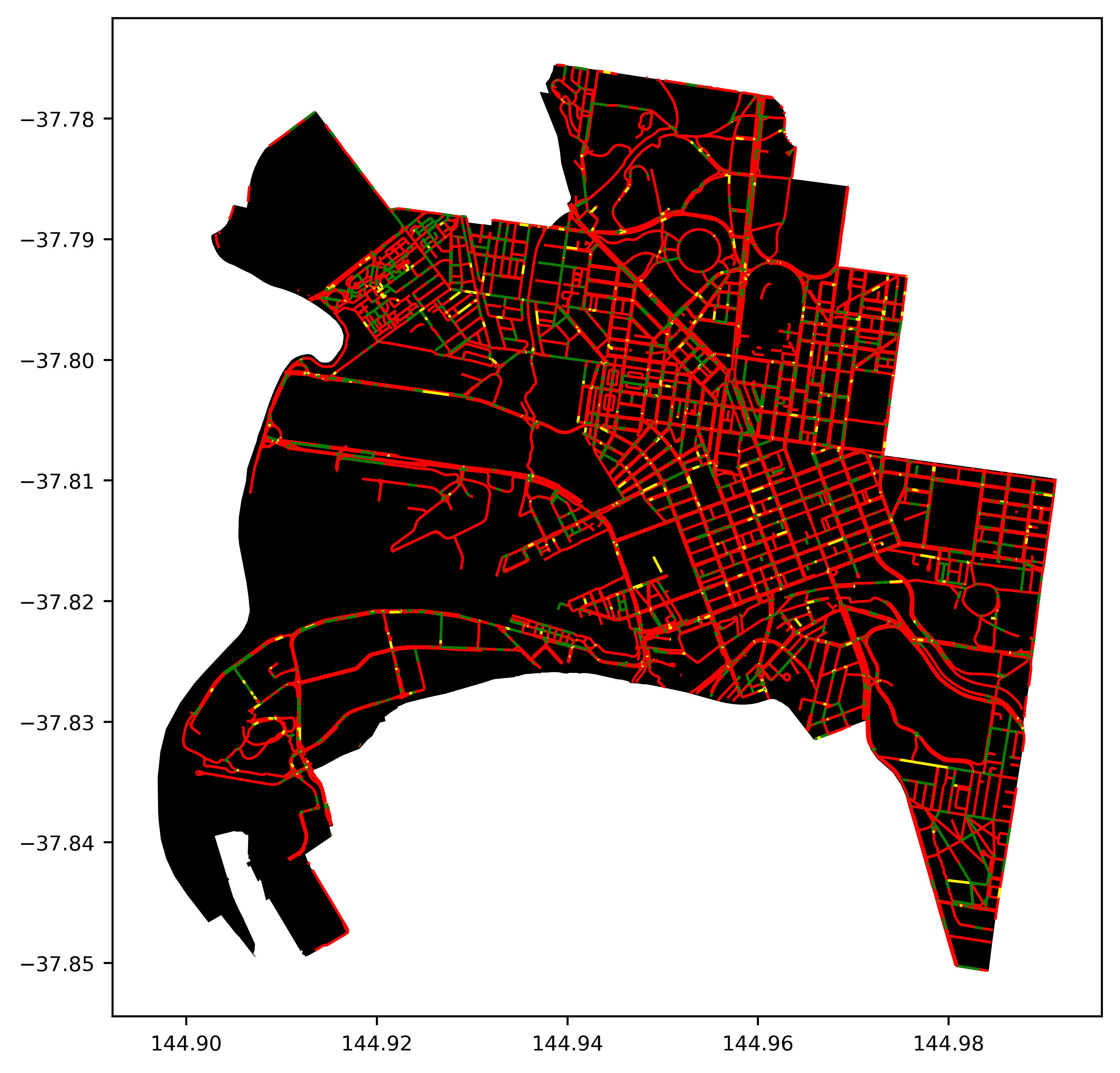}
        \subcaption{Sparsity - 0\%}
    \end{minipage}
    \begin{minipage}{0.31\textwidth}
        \centering
        \includegraphics[width=\textwidth]{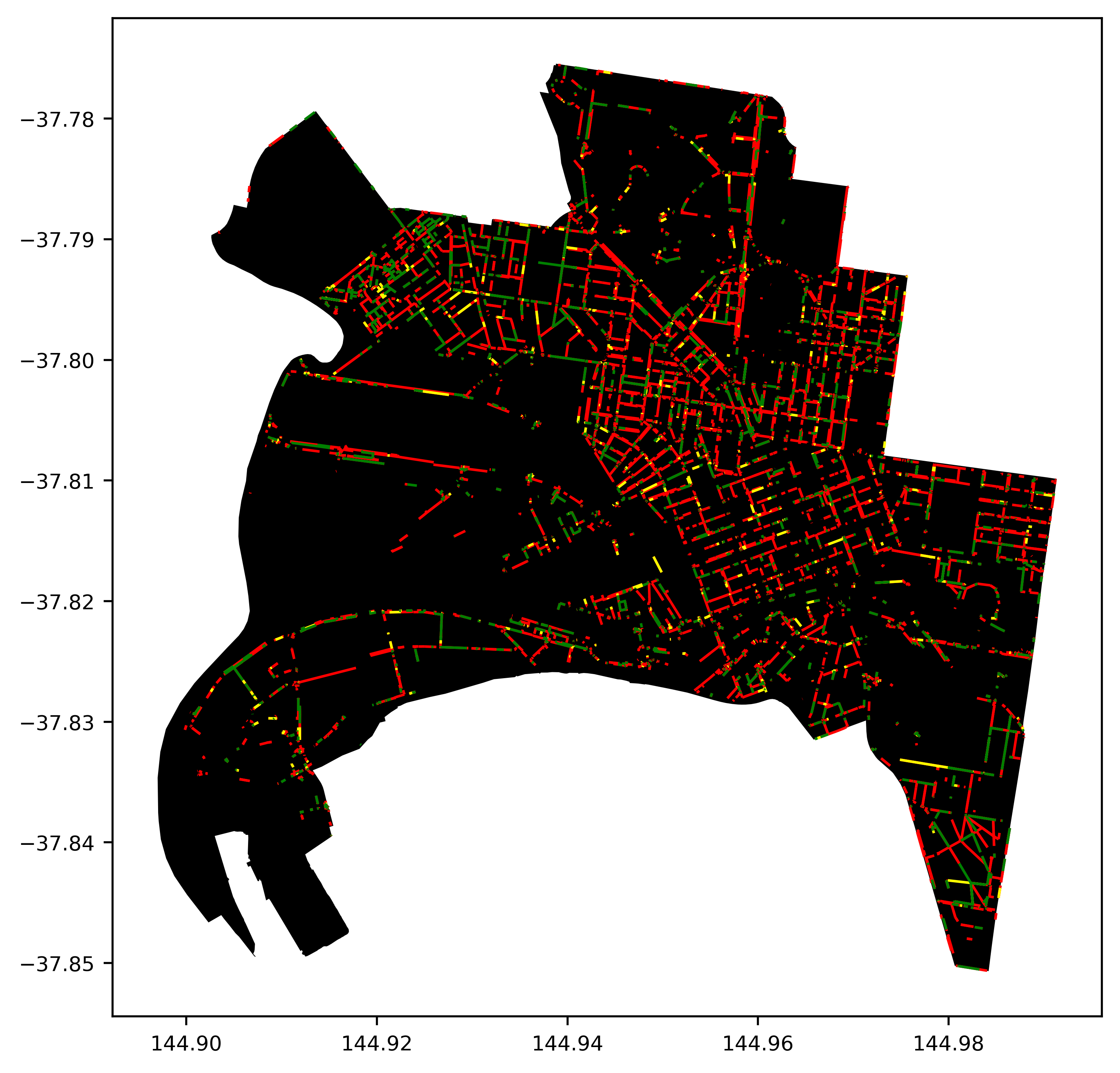}
        \subcaption{Sparsity - 20\%}
    \end{minipage}
    \begin{minipage}{0.31\textwidth}
        \centering
        \includegraphics[width=\textwidth]{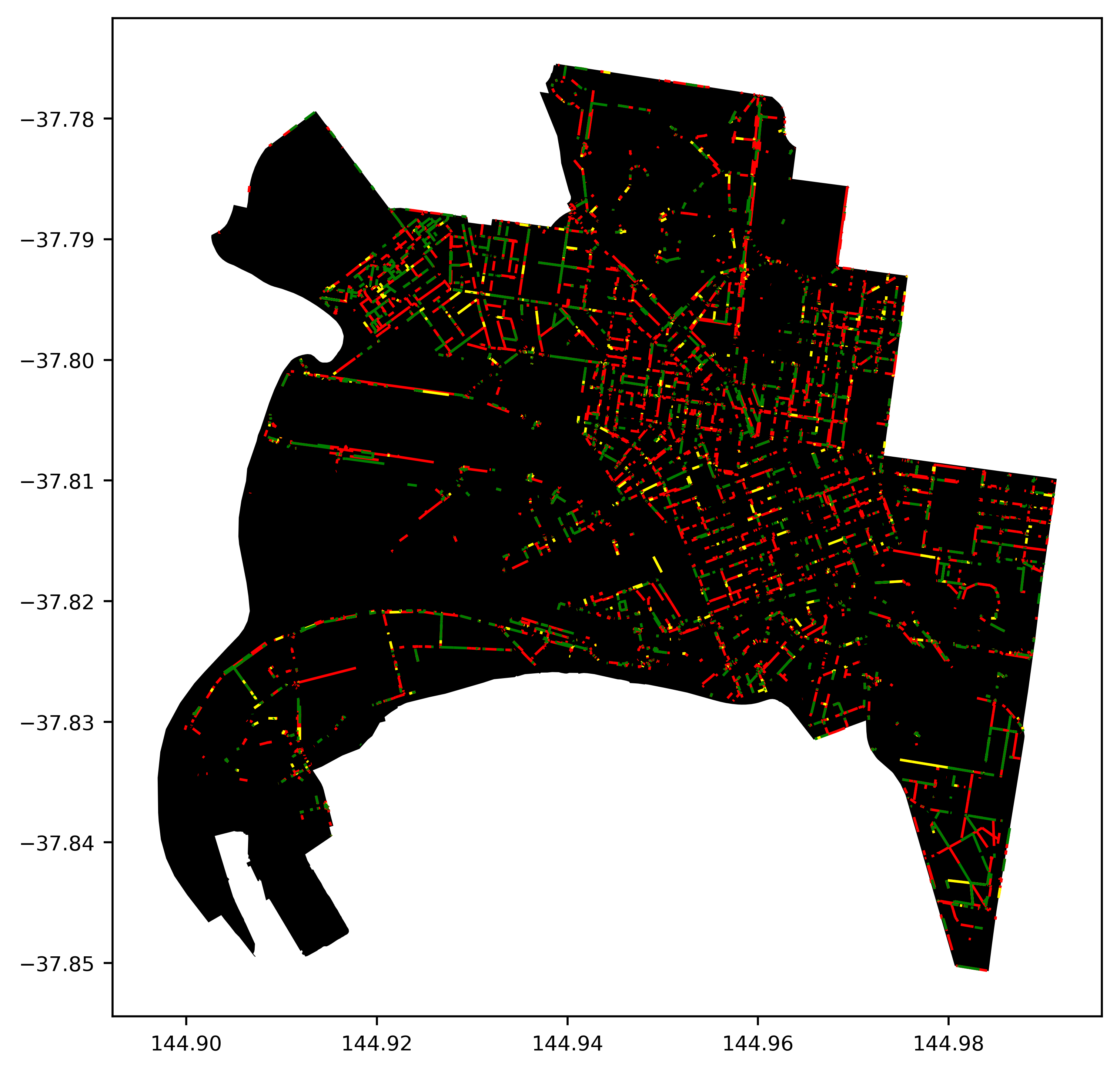}
        \subcaption{Sparsity - 50\%}
    \end{minipage}
    
    \vspace{2pt}
    
    \begin{minipage}{0.31\textwidth}
        \centering
        \includegraphics[width=\textwidth]{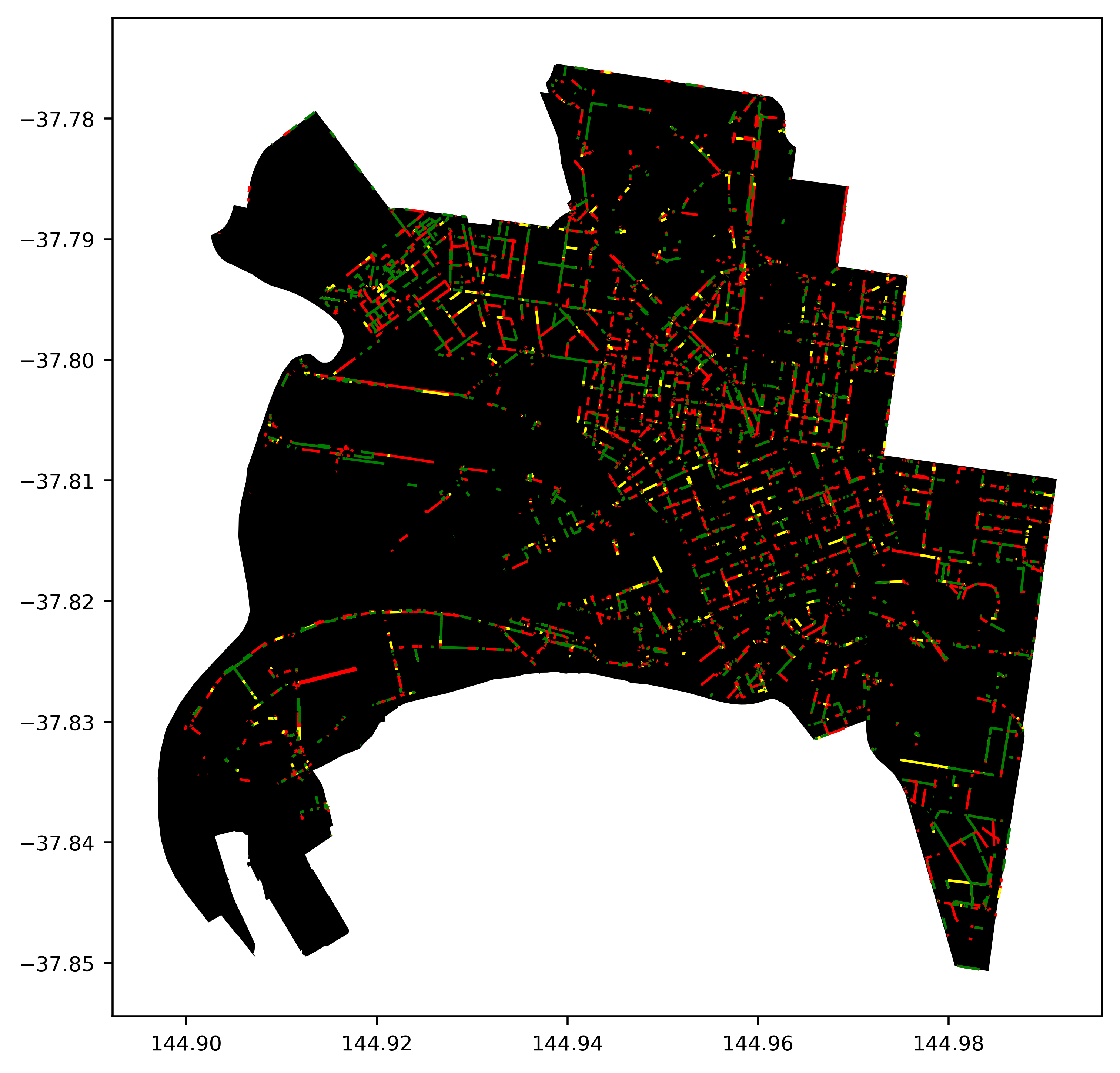}
        \subcaption{Sparsity - 60\%}
    \end{minipage}
    \begin{minipage}{0.31\textwidth}
        \centering
        \includegraphics[width=\textwidth]{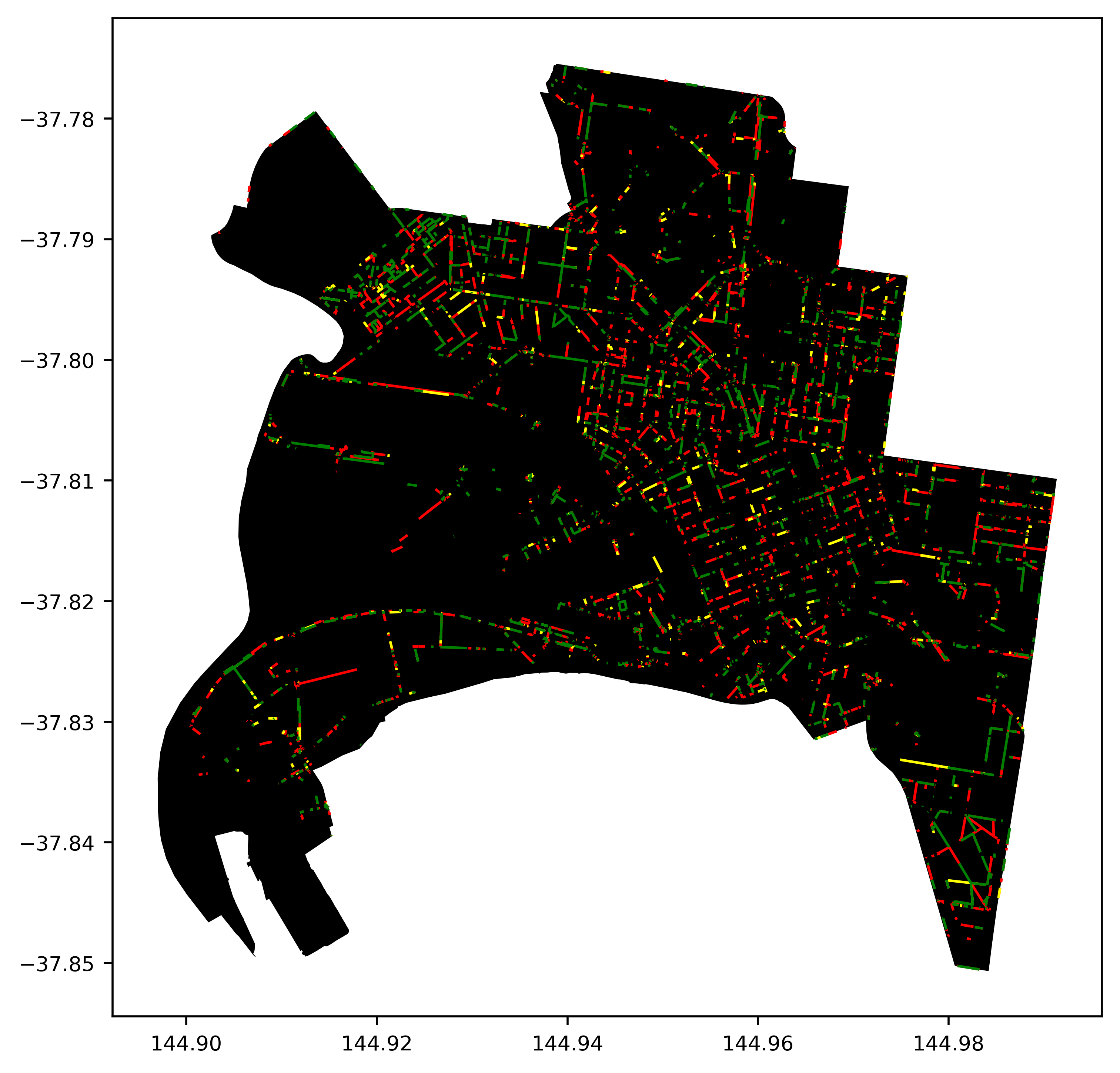}
        \subcaption{Sparsity - 70\%}
    \end{minipage}
    \begin{minipage}{0.31\textwidth}
        \centering
        \includegraphics[width=\textwidth]{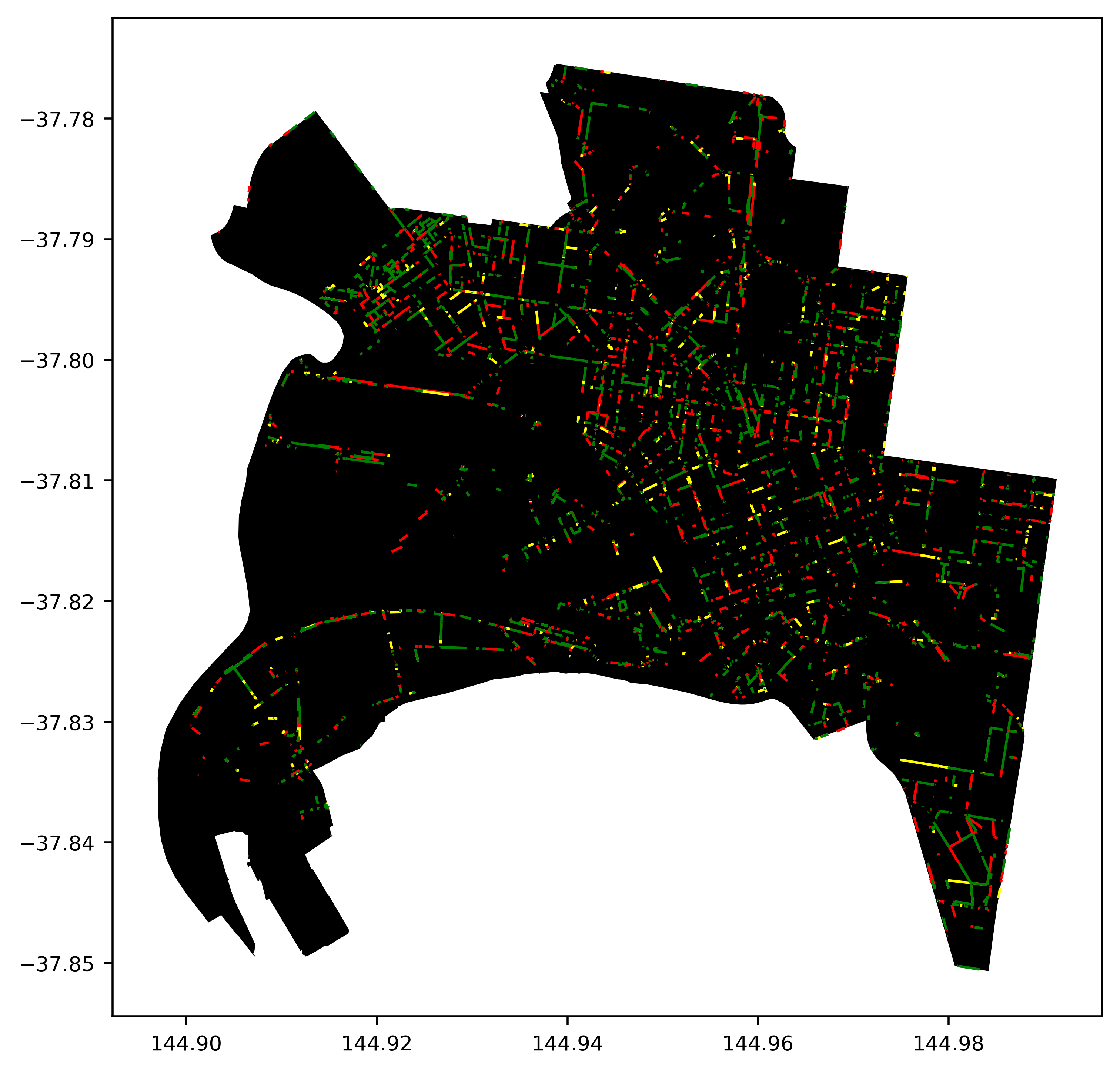}
        \subcaption{Sparsity - 80\%}
    \end{minipage}
    
    \vspace{2pt}
    
    \begin{minipage}{0.31\textwidth}
        \centering
        \includegraphics[width=\textwidth]{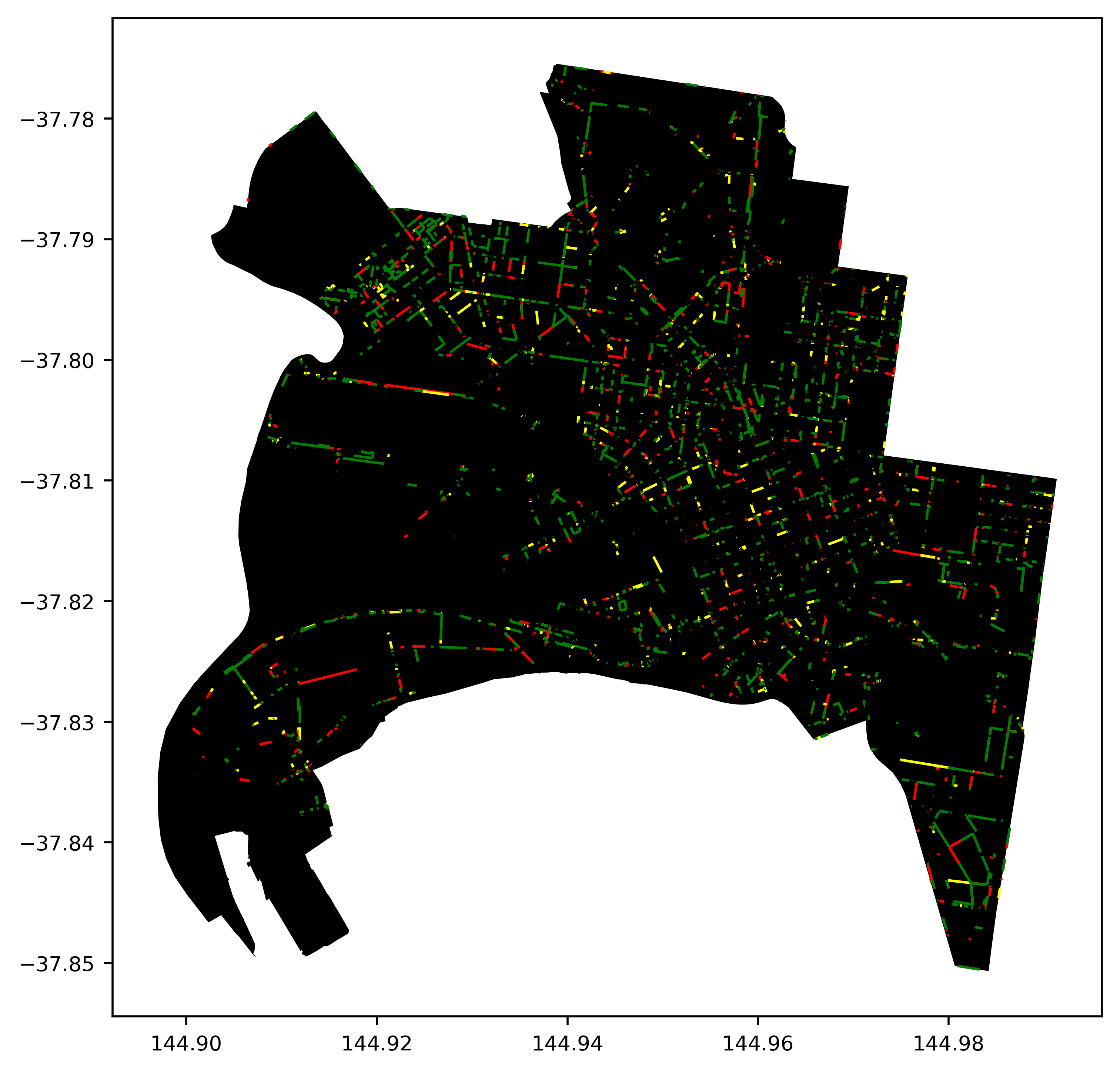}
        \subcaption{Sparsity - 90\%}
    \end{minipage}
    \begin{minipage}{0.31\textwidth}
        \centering
        \includegraphics[width=\textwidth]{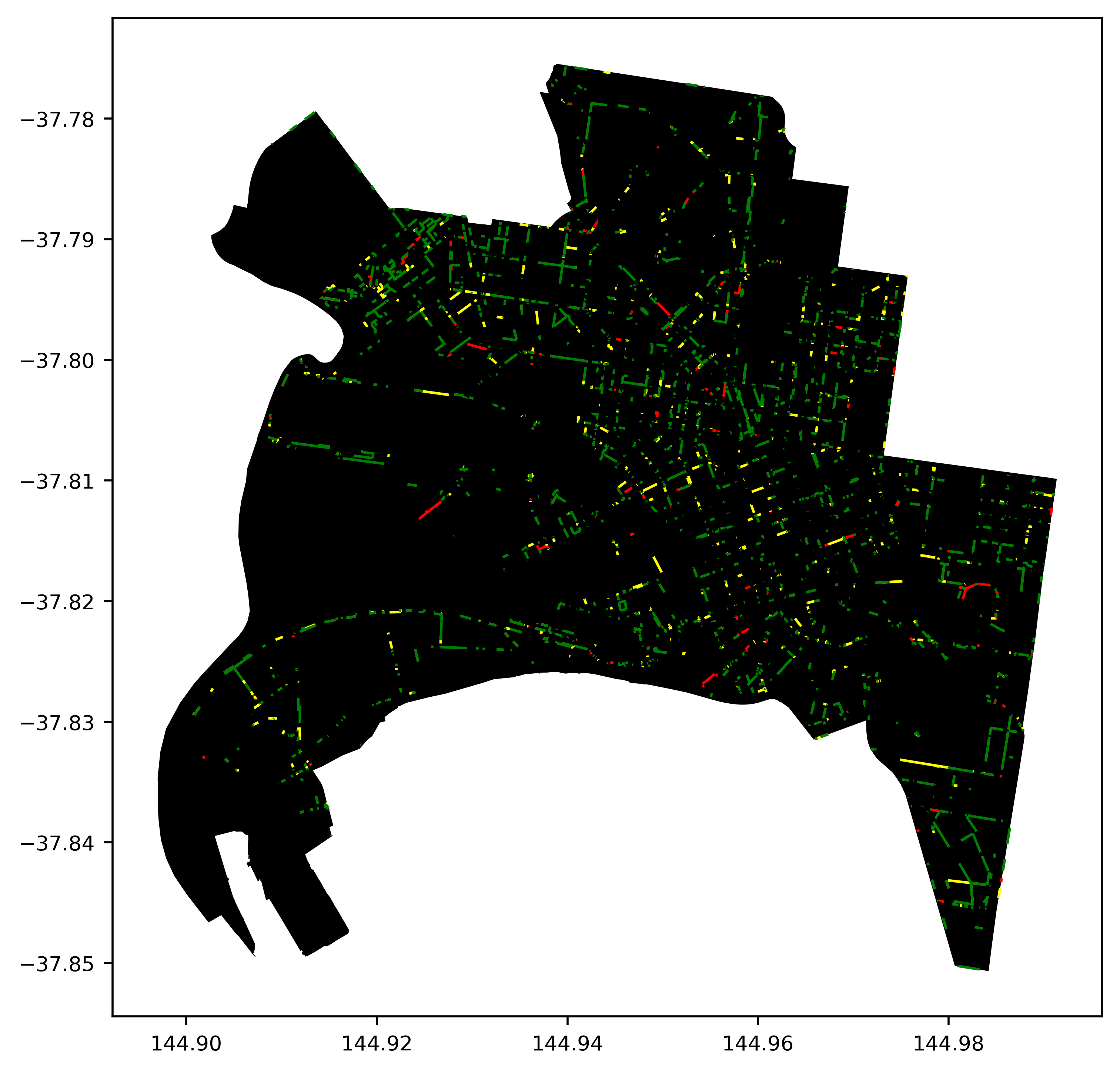}
        \subcaption{Sparsity - 99\%}
    \end{minipage}
    
    \label{fig:sparsity_simulation}
    \caption[]{Visualization of sparsity simulation for the bicycle network in the City of Melbourne. (a) Full bicycle network representation. (b–i) Graphs demonstrating different sparsity levels: 0\%, 20\%, 50\%, 60\%, 70\%, 80\%, 90\%, and 99\%. Red indicates training data, yellow indicates validation data, and green indicates testing data. As sparsity increases, fewer nodes are retained for training, simulating real-world scenarios of incomplete data. This visualization highlights the systematic reduction in training nodes.}
    \label{fig:sparsity_simulation}
    
\end{figure}

\section{Results and Discussion} \label{sec:results}
The results of this study provide a comprehensive evaluation of the Graph Convolutional Network (GCN) model performance in predicting link-level bicycling volumes within the City of Melbourne. 
We have systematically explored the impact of different GCN configurations, comparing them against traditional machine learning models under varying levels of data sparsity. 
The performance metrics including RMSE, MAE and MAPE offer a robust assessment of each model's accuracy and generalization capability. 
This section presents the findings from our experiments, highlighting the optimal GCN configurations, the comparative advantages of GCN over traditional models, and the critical role of data sparsity in influencing predictive performance. 

\subsection{Find the Optimal GCN configuration} \label{sec:gcn_config_result}
To identify the optimal Graph Convolutional Network (GCN) configuration for accurately predicting link-level bicycle volumes, we conducted a series of experiments with ten distinct configurations labeled A through J as already mentioned in \autoref{tab:confguration}. 
These configurations are designed to explore the impact of various architectural choices, including the number of GCN layers, the use of Batch Normalization, Dropout, Fully Connected Layers, and the inclusion of Residual connections. 

The evaluation is performed based on Root Mean Squared Error (RMSE) on the test dataset as shown in the  \autoref{tab:confguration_results} and it also focuses on comparing the performance with and without early stopping.
Our findings show that early stopping consistently improved model performance across most configurations by preventing overfitting during training. 
Configuration G emerged as the optimal architecture, achieving the lowest RMSE of 20.128 with early stopping enabled. 
This configuration includes a combination of three GCNConv layers, Batch Normalization, Dropout, and Fully Connected layers which striked the right balance between model complexity and regularization. 
Notably, it outperformed all the other configurations without excessively increasing the model's depth or number of parameters.

\begin{table}[H]
\centering
\caption{\small RMSE performance of different GCN configurations (as shown in Table \ref{tab:confguration}) with and without Early Stopping}
\label{tab:confguration_results}
\renewcommand{\arraystretch}{1.3}
\begin{tabular}{l|llllllllll}
\multicolumn{1}{c|}{\textbf{GCN Configuration}} & \multicolumn{1}{c}{\textbf{A}} & \multicolumn{1}{c}{\textbf{B}} & \multicolumn{1}{c}{\textbf{C}} & \multicolumn{1}{c}{\textbf{D}} & \multicolumn{1}{c}{\textbf{E}} & \multicolumn{1}{c}{\textbf{F}} & \multicolumn{1}{c}{\textbf{G}} & \multicolumn{1}{c}{\textbf{H}} & \multicolumn{1}{c}{\textbf{I}} & \multicolumn{1}{c}{\textbf{J}} \\ \hline
\textbf{\begin{tabular}[c]{@{}l@{}}RMSE without \\ Early Stopping\end{tabular}} & 23.255 & 21.039 & 22.417 & 20.622 & 22.115 & 20.975 & 20.334 & 21.494 & 20.755 & 20.373 \\ \hline
\textbf{\begin{tabular}[c]{@{}l@{}}RMSE with \\ Early Stopping\end{tabular}} & 23.008 & 20.812 & 22.292 & 20.2 & 22.048 & 20.034 & \textbf{20.128} & 21.488 & 20.6 & 20.163
\end{tabular}
\end{table}

Configurations A to D represented relatively simpler architectures, consisting of two or three GCNConv layers with fewer feature channels (32 and 64). 
These configurations were intended to establish a performance baseline. 
While these models demonstrated reasonable performance, they failed to capture the intricate spatial dependencies in the bicycling network as effectively as the more complex architectures. 
For instance, configuration A achieved an RMSE of 23.008 with early stopping, which was notably higher compared to the more sophisticated configurations.

The deeper architectures such as configurations G, I, and J, leveraged additional GCNConv layers and an increased number of feature channels (up to 256) to capture more detailed and nuanced relationships within the graph. 
The gradual introduction of architectural enhancements, such as Batch Normalization and Dropout, helped to improve model generalization. 
However despite being more complex, configurations H and J could not surpass configuration G in performance, likely due to overfitting resulting from the increased depth and number of parameters.

Overall, the experiments demonstrate that while adding depth and complexity to the GCN architecture can enhance its ability to learn from spatially structured data, there is a critical threshold beyond which performance gains plateau or even diminish. 
The results clearly indicate that Configuration G strikes an optimal balance between model depth, feature channels, and regularization techniques and highlight the importance of designing a balanced architecture to effectively leverage the strengths of GCNs for link-level bicycling volume prediction, where the spatial relationships are complex.
Based on these findings, configuration G is selected for further experiments, including the evaluation of the model's robustness to varying levels of data sparsity and its comparison with traditional machine learning models.

\subsection{Find the optimal ML parameters}
To determine the optimal hyperparameters for each classical machine learning model - LR, SVM and RF, we conducted an extensive hyperparameter tuning process using grid search combined with cross-validation. 
The goal was to identify the most effective configuration for each model to predict link-level bicycling volumes in the City of Melbourne. \autoref{tab:ml_confguration_results} summarizes the optimal hyperparameters obtained along with the corresponding performance metrics: RMSE, MAE, MAPE.

The optimal configuration for LR was found using L2 regularization with a regularization strength ($\alpha$) of 0.1. This setup yielded an RMSE of 26.108, MAE of 11.625, and MAPE of 3.159.
The results indicate that while LR is relatively easy to configure, its performance is limited when it comes to capturing complex relationships, especially in a non-linear and graph-based dataset like bicycling volumes. 
For the SVM model, we tested different kernel types (linear, poly, rbf, sigmoid) along with the regularization parameter (C) and kernel coefficient ($\gamma$). 
The optimal hyperparameters for SVM were identified as an RBF kernel, a regularization parameter (C) of 10, and a kernel coefficient ($\gamma$) of 0.01. 
With this configuration, the SVM model achieved an RMSE of 29.854, MAE of 10.029, and MAPE of 1.97. Although SVM produced a lower MAPE compared to LR, its RMSE and MAE suggest that it struggles with absolute error minimization.
For the random forest model, we experimented with hyperparameters such as the number of estimators, maximum tree depth, minimum samples to split, and minimum samples at each leaf node. 
The best configuration was found to include 400 estimators, a maximum depth of 20, a minimum split of 2 samples, and 1 sample per leaf node. 
Under these settings, RF model produced an RMSE of 23.848, MAE of 10.889, and MAPE of 2.726. 
RF demonstrated the best overall performance among traditional machine learning models, reflecting its ability to handle non-linear relationships and complex patterns in the dataset effectively.

\begin{table}[H]
\centering
\caption{\small Performance of Classical Machine Learning Algorithms with optimally selected Hyper-parameters (HP)}
\label{tab:ml_confguration_results}
\renewcommand{\arraystretch}{1.3}
\begin{tabular}{l|lll|ccc}
\textbf{Method}      & \multicolumn{1}{c}{\textbf{HP Name}} & \multicolumn{1}{c}{\textbf{HP Range}}                                         & \multicolumn{1}{c|}{\textbf{Optimal HP}} & \textbf{RMSE}           & \textbf{MAE}            & \textbf{MAPE}          \\ \hline
\multirow{2}{*}{LR}  & Regularization Strength ($\alpha$)          & {[}0.0001, 1{]}                                                               & 0.1                                      & \multirow{2}{*}{26.108} & \multirow{2}{*}{11.625} & \multirow{2}{*}{3.159} \\
                     & Regularization Type                  & \begin{tabular}[c]{@{}l@{}}L1 (Lasso), L2 (Ridge), \\ ElasticNet\end{tabular} & L2                                       &                         &                         &                        \\ \hline
\multirow{3}{*}{SVM} & C (Regularization Parameter)         & {[}0.1, 100{]}                                                                & 10                                       & \multirow{3}{*}{29.854} & \multirow{3}{*}{10.029} & \multirow{3}{*}{1.97}  \\
                     & Gamma (Kernel Coefficient)           & {[}0.001, 1{]}                                                                & 0.01                                     &                         &                         &                        \\
                     & Kernel Type                          & \begin{tabular}[c]{@{}l@{}}{[}linear, poly, rbf, \\ sigmoid{]}\end{tabular}   & rbf                                      &                         &                         &                        \\ \hline
\multirow{4}{*}{RF}  & Number of Estimators                 & {[}100, 1000{]}                                                               & 400                                      & \multirow{4}{*}{23.848} & \multirow{4}{*}{10.889} & \multirow{4}{*}{2.726} \\
                     & Max Depth                            & {[}3, 10{]}                                                                   & 20                                       &                         &                         &                        \\
                     & Min Samples Split                    & {[}2, 10{]}                                                                   & 2                                        &                         &                         &                        \\
                     & Min Samples Leaf                     & {[}1, 5{]}                                                                    & 1                                        &                         &                         &                        \\ \hline
\end{tabular}
\end{table}

Overall, the optimal hyperparameter tuning results as presented in \autoref{tab:ml_confguration_results}, demonstrate that Random Forest outperformed both Linear Regression and SVM in terms of RMSE and MAE, indicating a better fit to the data and greater ability to minimize prediction errors. 
However, SVM achieved a lower MAPE, suggesting it was more effective in percentage-based error metrics, which may be beneficial for assessing relative prediction accuracy. 
Overall, these traditional machine learning models provided useful performance benchmarks, but their limitations in capturing the intricate spatial dependencies of bicycling volumes are evident when compared to more advanced models, such as the Graph Convolutional Network (GCN) discussed in the next section.

\subsection{Effects of data sparsity on the models performance}

\begin{figure}[H]
    \centering
    \includegraphics[width=1\linewidth]{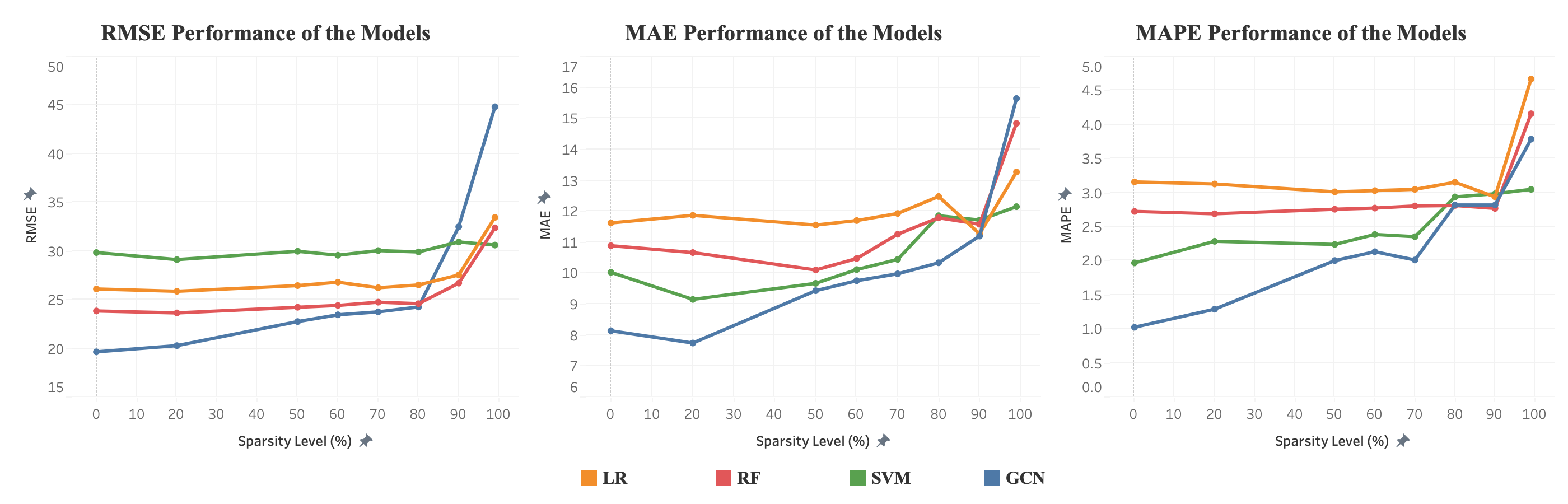}
    \caption{\small Performance of GCN and Traditional Machine Learning Models Across Different Levels of Data Sparsity}
    \label{fig:model_performance_and_sparsity}
\end{figure}

This section presents a detailed evaluation of how varying levels of data sparsity affect the performance of the Graph Convolutional Network (GCN) compared to traditional machine learning models LR, SVM, RF — in estimating link-level Annual Average Daily Bicycle (AADB) counts. The results of our experiments under different sparsity levels, from 0\% to 99\%, are summarized in Table \ref{tab:sparsity_results}, and the corresponding performance metrics are visualized in Figure \ref{fig:model_performance_and_sparsity}.

As shown in Table \ref{tab:sparsity_results}, at 0\% sparsity, where the dataset is fully populated with data for 12,746 nodes, the GCN model outperforms traditional models, achieving the lowest RMSE (19.644), MAE (8.140), and MAPE (1.025). 
This highlights the GCN’s ability to effectively capture the spatial dependencies inherent in the bicycle network which ML models lack as they fail to explicitly account for the underlying graph structures.

\textbf{Low to Moderate Sparsity (20\% - 60\%):}
As data sparsity increases to 20\% and further to 60\%, GCN consistently outperforms LR, SVM, and RF across all evaluation metrics. 
At 20\% sparsity, the GCN achieved an RMSE of 20.284, which represents only a marginal increase from its RMSE at 0\% sparsity. 
In contrast, SVM and RF show more considerable performance degradation, highlighting their struggles to adapt to missing data. 
This trend indicates that the GCN's graph-based approach can continue leveraging the available spatial relationships, even when some data points are removed.

\textbf{Moderate to High Sparsity (70\% - 80\%):}
At higher levels of sparsity, such as 70\% and 80\%, GCN continues to maintain superior performance compared to traditional models, with an RMSE of 23.757 and 24.258, respectively. 
Although the GCN's performance declines at these levels, it still benefits from its ability to aggregate features from neighboring nodes, which mitigates the effect of missing data points. 
The SVM model, while also exhibiting declining performance, showed a relatively smaller change compared to LR, indicating its capability in handling some level of data sparsity.

\begin{table}[H]
\centering
\caption{\small Performance Comparison of GCN and Machine Learning Algorithms at Different Sparsity Levels}
\label{tab:sparsity_results}
\renewcommand{\arraystretch}{1.2}
\begin{tabular}{ccl|lll|lll}
\multicolumn{1}{l}{\multirow{2}{*}{\textbf{\begin{tabular}[c]{@{}l@{}}Sparsity\\ Level (\%)\end{tabular}}}} &
  \multirow{2}{*}{\textbf{\begin{tabular}[c]{@{}c@{}}No. of labelled nodes \\ for model training\end{tabular}}} &
   &
  \multicolumn{3}{c|}{\textbf{Testing}} &
  \multicolumn{3}{c}{\textbf{Validation}} \\ \cline{3-9} 
\multicolumn{1}{l}{} &
   &
  \textbf{Method} &
  \textbf{RMSE} &
  \textbf{MAE} &
  \textbf{MAPE} &
  \textbf{RMSE} &
  \textbf{MAE} &
  \textbf{MAPE} \\ \hline
\multirow{4}{*}{0}     & \multirow{4}{*}{12746} & LR  & 26.108 & 11.625 & 3.159 & 26.904 & 12.459 & 2.837 \\
                       &                        & SVM & 29.854 & 10.029 & 1.970 & 31.197 & 10.242 & 2.010 \\
                       &                        & RF  & 23.848 & 10.889 & 2.726 & 23.461 & 10.298 & 2.530 \\
                       &                        & GCN & 19.644 & 8.140  & 1.025 & 21.124 & 8.045  & 0.960 \\ \hline
\multirow{4}{*}{20}    & \multirow{4}{*}{10196} & LR  & 25.872 & 11.871 & 3.129 & 26.685 & 11.876 & 2.805 \\
                       &                        & SVM & 29.130 & 9.154  & 2.287 & 31.790 & 10.521 & 2.144 \\
                       &                        & RF  & 23.647 & 10.668 & 2.691 & 24.056 & 10.627 & 2.531 \\
                       &                        & GCN & 20.284 & 7.745  & 1.290 & 19.937 & 7.802  & 0.959 \\ \hline
\multirow{4}{*}{50}    & \multirow{4}{*}{6373}  & LR  & 26.456 & 11.556 & 3.013 & 26.650 & 12.709 & 2.737 \\
                       &                        & SVM & 29.983 & 9.674  & 2.241 & 31.828 & 10.183 & 2.099 \\
                       &                        & RF  & 24.228 & 10.106 & 2.758 & 24.158 & 10.741 & 2.561 \\
                       &                        & GCN & 22.758 & 9.436  & 2.004 & 21.919 & 9.357  & 0.953 \\ \hline
\multirow{4}{*}{60}    & \multirow{4}{*}{5098}  & LR  & 26.806 & 11.702 & 3.031 & 27.429 & 11.893 & 2.742 \\
                       &                        & SVM & 29.578 & 9.912  & 2.387 & 31.117 & 10.040 & 2.122 \\
                       &                        & RF  & 24.416 & 10.475 & 2.775 & 23.597 & 10.616 & 2.547 \\
                       &                        & GCN & 23.454 & 9.758  & 2.134 & 23.225 & 9.045  & 0.946 \\ \hline
\multirow{4}{*}{70}    & \multirow{4}{*}{3823}  & LR  & 26.237 & 11.930 & 3.050 & 27.526 & 12.581 & 2.764 \\
                       &                        & SVM & 30.054 & 9.446  & 2.355 & 31.801 & 10.898 & 2.436 \\
                       &                        & RF  & 24.751 & 11.260 & 2.806 & 25.043 & 11.171 & 2.623 \\
                       &                        & GCN & 23.757 & 10.979 & 2.013 & 23.369 & 10.182 & 0.960 \\ \hline
\multirow{4}{*}{80}    & \multirow{4}{*}{2549}  & LR  & 26.522 & 12.479 & 3.154 & 26.997 & 12.572 & 2.812 \\
                       &                        & SVM & 29.916 & 10.337 & 2.939 & 31.730 & 10.324 & 3.541 \\
                       &                        & RF  & 24.594 & 10.791 & 2.813 & 25.433 & 11.588 & 2.657 \\
                       &                        & GCN & 24.258 & 11.858 & 2.820 & 33.137 & 11.952 & 0.919 \\ \hline
\multirow{4}{*}{90}    & \multirow{4}{*}{1274}  & LR  & 27.549 & 11.272 & 2.939 & 29.748 & 12.772 & 2.732 \\
                       &                        & SVM & 30.941 & 9.718  & 2.985 & 32.186 & 11.293 & 3.541 \\
                       &                        & RF  & 26.707 & 10.583 & 2.770 & 27.249 & 11.539 & 2.649 \\
                       &                        & GCN & 32.507 & 11.202 & 2.820 & 70.181 & 11.480 & 0.953 \\ \hline
\multirow{4}{*}{99} & \multirow{4}{*}{127}   & LR  & 33.469 & 13.271 & 4.670 & 35.578 & 12.949 & 4.323 \\
                       &                        & SVM & 30.622 & 10.150 & 3.050 & 32.779 & 11.156 & 3.033 \\
                       &                        & RF  & 32.392 & 14.843 & 4.160 & 33.398 & 15.621 & 4.281 \\
                       &                        & GCN & 44.838 & 15.644 & 3.787 & 48.889 & 19.329 & 3.820 \\ \hline
\end{tabular}
\end{table}

\textbf{Extreme Sparsity Levels (90\% - 99\%):}
At 90\% sparsity, with 1,274 nodes for training, the GCN performance starts to significantly degrade, with an RMSE of 32.507 suggesting that the effectiveness of GCN diminishes when data becomes excessively sparse. 
RF exhibits a more gradual decline, achieving an RMSE of 26.707.
The reduced complexity of RF allows it to handle extreme sparsity more reliably, though with reduced accuracy compared to GCN at lower sparsity levels. 
At extreme sparsity levels (99\%), the labeled nodes drop to only 127 which is similar to the real-world scenario of count locations in the City of Melbourne. 
At this level, all models suffer from high RMSE, MAE, and MAPE values, but GCN's performance falls more drastically, with RMSE reaching 44.838. 
The RF model, though still not ideal, becomes relatively more reliable under these conditions, with an RMSE of 32.392. 
The inability of GCN to rely on spatial aggregation due to a lack of sufficient labeled nodes results in its significant decline.

From Figure \ref{fig:model_performance_and_sparsity}, which illustrates the RMSE, MAE, and MAPE metrics across sparsity levels, it is evident that while GCN is the best model under low to moderate sparsity conditions, it struggles under extreme sparsity, especially beyond 80\%. 
It also emphasize the need for adequate count data to fully leverage the strengths of GCNs.
Also, the relatively better performance of traditional models under extreme sparsity should not be interpreted as these models being better equipped to handle data sparsity in a broader sense. 
Instead, it highlights the trade-off between model complexity and data availability. Traditional models perform adequately in sparse conditions because they are less reliant on complex data structures, whereas GCNs require a sufficient density of data to fully leverage their strengths in modeling spatial relationships \citep{khemani2024review}. 
Therefore, in more typical scenarios where data is not excessively sparse, GCNs offer significant advantages, especially when modeling complex, graph-structured data like bicycling networks.
In real-world applications where data sparsity is a significant issue, these findings suggest that while GCNs offer substantial benefits especially in their ability to model complex spatial relationships, there is also a need to develop strategies to mitigate the impact of extreme data sparsity.

\section{Conclusion and Future Research Directions} \label{sec:conclusion}
This study addresses the unique challenges of link-level bicycling volume estimation by leveraging a Graph Convolutional Network (GCN) architecture specifically designed for the geographical area of the City of Melbourne. 
While traditional link-level volume estimation models have demonstrated effectiveness in motorized traffic applications, they fall short in the bicycling context due to the sparse nature of the available data and the complex mobility patterns unique to bicycling. 
To the best of our knowledge, this is the first study to evaluate the effects of data sparsity and utilize a Graph Convolutional Network (GCN) architecture specifically to model link-level bicycling volumes in a real-world urban context, thereby filling a significant gap in the existing literature.

By integrating OpenStreetMap (OSM) network data with Strava Metro’s bicycling activity data, we simulate varying levels of data sparsity and evaluate the robustness of the GCN model under these conditions. 
Our findings reveal that the GCN outperforms traditional machine learning models particularly in scenarios with low to moderate data sparsity. 
Specifically, the optimal GCN configuration achieved the lowest Root Mean Squared Error (RMSE) across different sparsity levels, underscoring its superior ability to capture spatial dependencies inherent in bicycling networks.

However, the study also highlights the limitations of GCNs in environments with extreme data sparsity. 
As sparsity levels exceeded 70\%, the GCN's performance degraded significantly, more so than the traditional models, which exhibited a more gradual decline. 
This suggests that while GCNs are highly effective with sufficient data, their performance is compromised when data becomes extremely sparse, as is often the case in many urban areas, including Melbourne.
The inherent biases in Strava Metro data, which tends to over-represent recreational cyclists and underreport quieter routes, were acknowledged as potential limitations that could skew the model’s predictions. 
Despite these biases, Strava data remains invaluable for its extensive spatial coverage, providing a robust foundation for simulating data sparsity and evaluating model performance.

The implications of this research are significant for urban planners and policymakers aiming to enhance bicycling infrastructure. 
GCNs offer a promising tool for accurately modeling and predicting bicycling volumes, facilitating informed decision-making to promote sustainable transportation. 
Nonetheless, the decline in GCN performance under high sparsity underscores the necessity for strategies to mitigate data sparsity, such as integrating additional data sources like real-time traffic conditions or environmental factors.

Future research should prioritize enhancing the robustness of Graph Convolutional Networks (GCNs) in environments characterized by extreme data sparsity. One promising approach is the development of hybrid models that combine GCNs with other machine learning techniques, leveraging their complementary strengths to improve performance under high sparsity conditions. Additionally, the incorporation of auxiliary data sources—such as real-time traffic sensors, environmental factors, and demographic information—can significantly enrich the feature set, compensating for missing data and enhancing the model's predictive capabilities.

Exploring advanced data augmentation methods is another critical avenue, as sophisticated techniques can artificially increase data density and provide more diverse training samples, thereby improving model generalization. Moreover, expanding the study to ensure generalizability across diverse urban contexts is essential. Applying the proposed GCN approach to other cities with varied bicycling networks will validate its adaptability and effectiveness in different settings, ensuring broader applicability.

Addressing and mitigating data biases inherent in crowd-sourced data, such as those from Strava, is also crucial. Developing methods to correct these biases will ensure more accurate and representative model predictions, enhancing the reliability of bicycling volume estimates. Furthermore, integrating temporal dynamics into the model by incorporating temporal layers or time-series data will enhance its ability to capture temporal patterns, allowing for more dynamic and responsive predictions that reflect fluctuating bicycling demand over time.

Lastly, extending the model to support real-time prediction capabilities represents a significant advancement. Real-time bicycling volume predictions can be instrumental for immediate infrastructure adjustments and traffic management, providing urban planners and policymakers with timely insights to promote sustainable transportation. By pursuing these research directions, future studies can further refine and extend the applicability of GCNs in bicycling volume estimation, making them even more valuable for urban mobility planning and sustainable transportation initiatives.

In conclusion, this study demonstrates the substantial potential of Graph Convolutional Networks in advancing link-level bicycling volume estimation, particularly in data-rich environments. 
As cities continue to prioritize sustainable transportation, the integration of GCNs into urban mobility planning presents a valuable opportunity to harness complex spatial relationships and drive informed infrastructure investments.


\subsection*{Acknowledgements}
The CYCLED (CitY-wide biCycLing Exposure modelling) Study is funded by an Australian Research Council Discovery Project (DP210102089).
Mohit Gupta's PhD scholarship and Debjit Bhowmick were supported by the Australian Research Council Discovery Project (DP210102089).
Ben Beck was supported by an Australian Research Council Future Fellowship (FT210100183). This work includes aggregated and de-identified data from Strava Metro \citep{StravaMetro}.

\bibliographystyle{apalike}
\bibliography{main.bib}

\clearpage

\appendix
\section{Appendix}
\subsection{Training and Validation Loss Curves for GCN Configurations A-J}
\begin{figure}[H]
    \centering
    \begin{minipage}{0.31\textwidth}
        \centering
        \includegraphics[width=\textwidth]{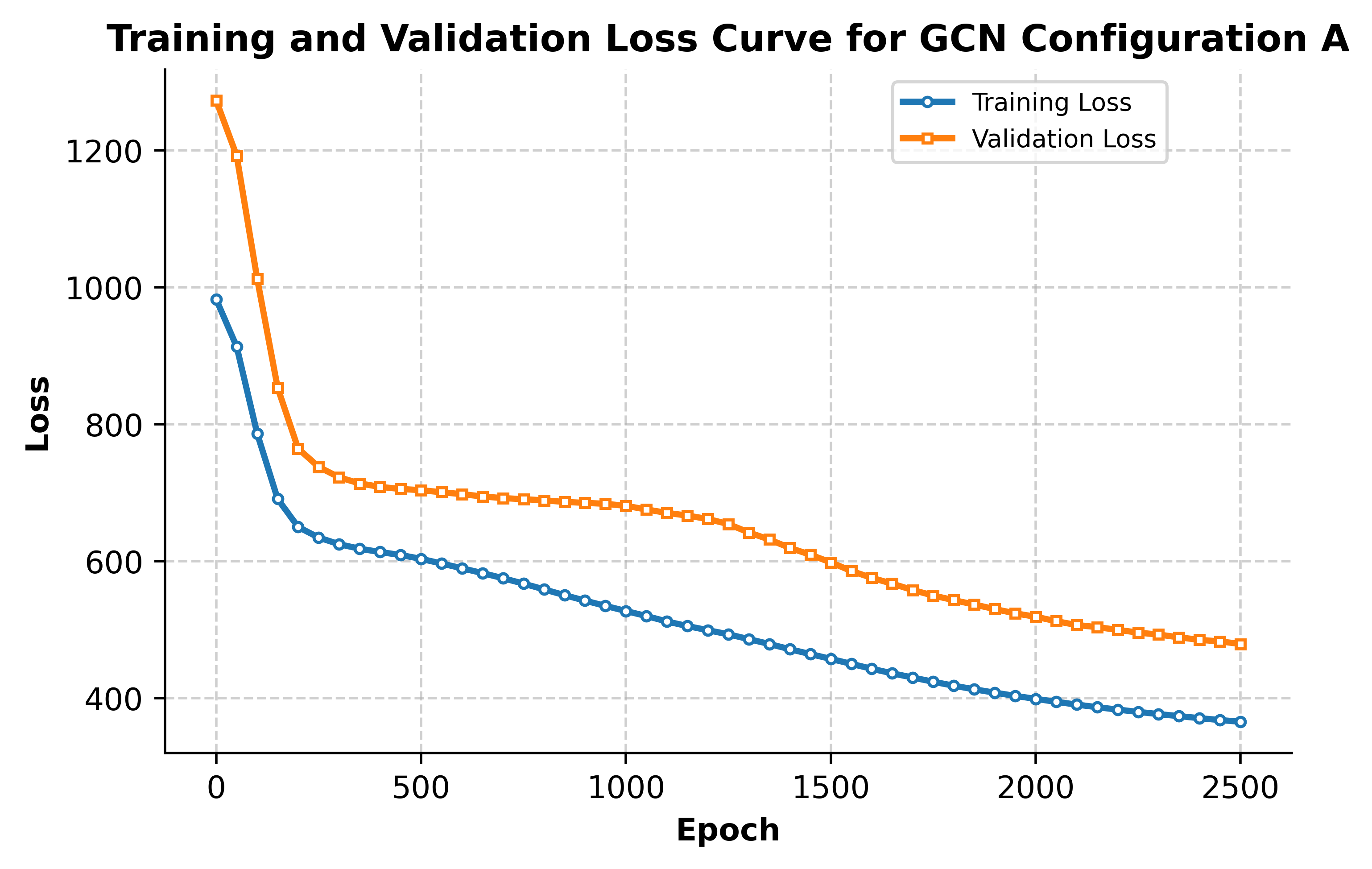}
        \subcaption{GCN Configuration A}
    \end{minipage}
    \begin{minipage}{0.31\textwidth}
        \centering
        \includegraphics[width=\textwidth]{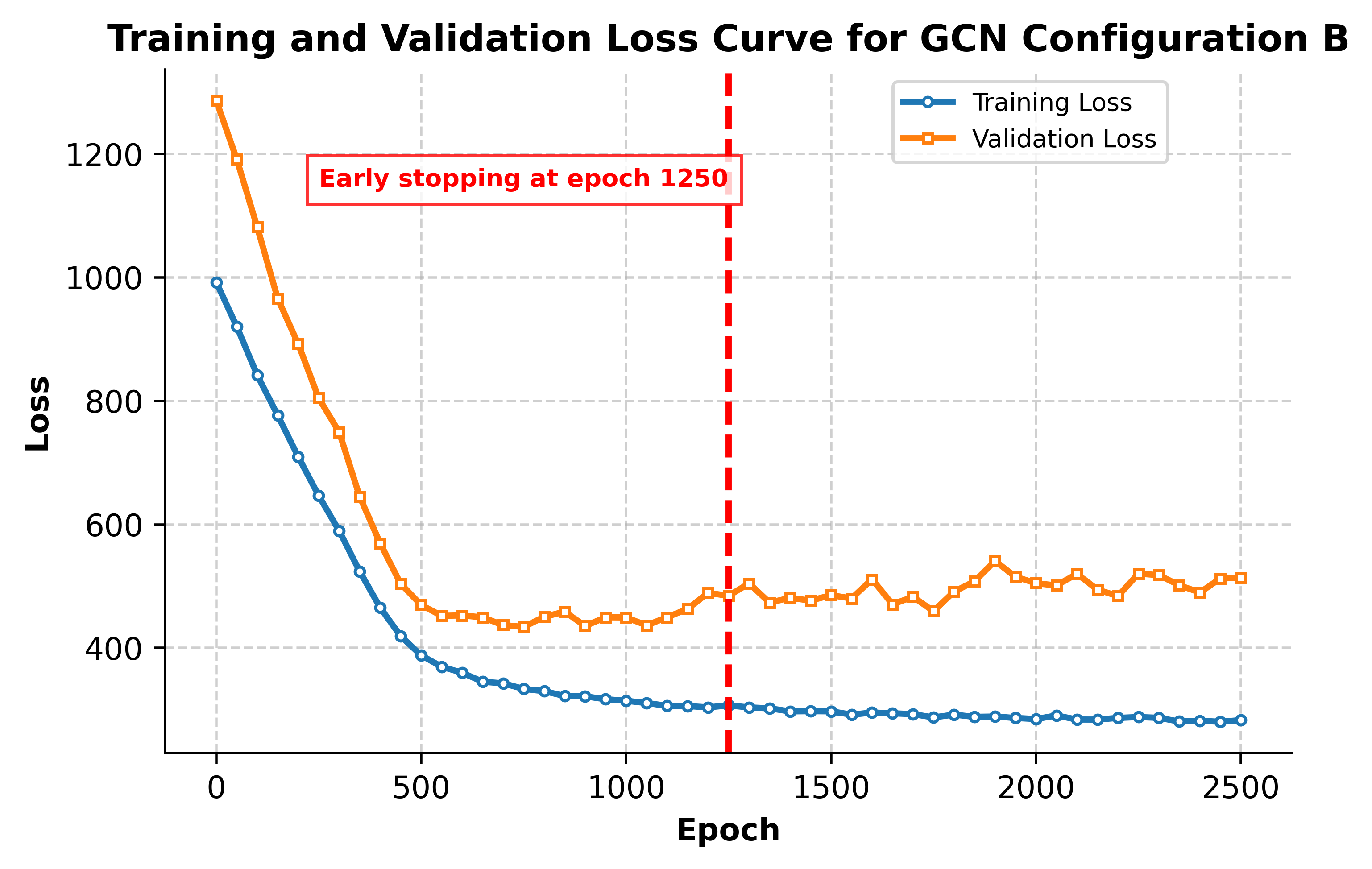}
        \subcaption{GCN Configuration B}
    \end{minipage}
    \begin{minipage}{0.31\textwidth}
        \centering
        \includegraphics[width=\textwidth]{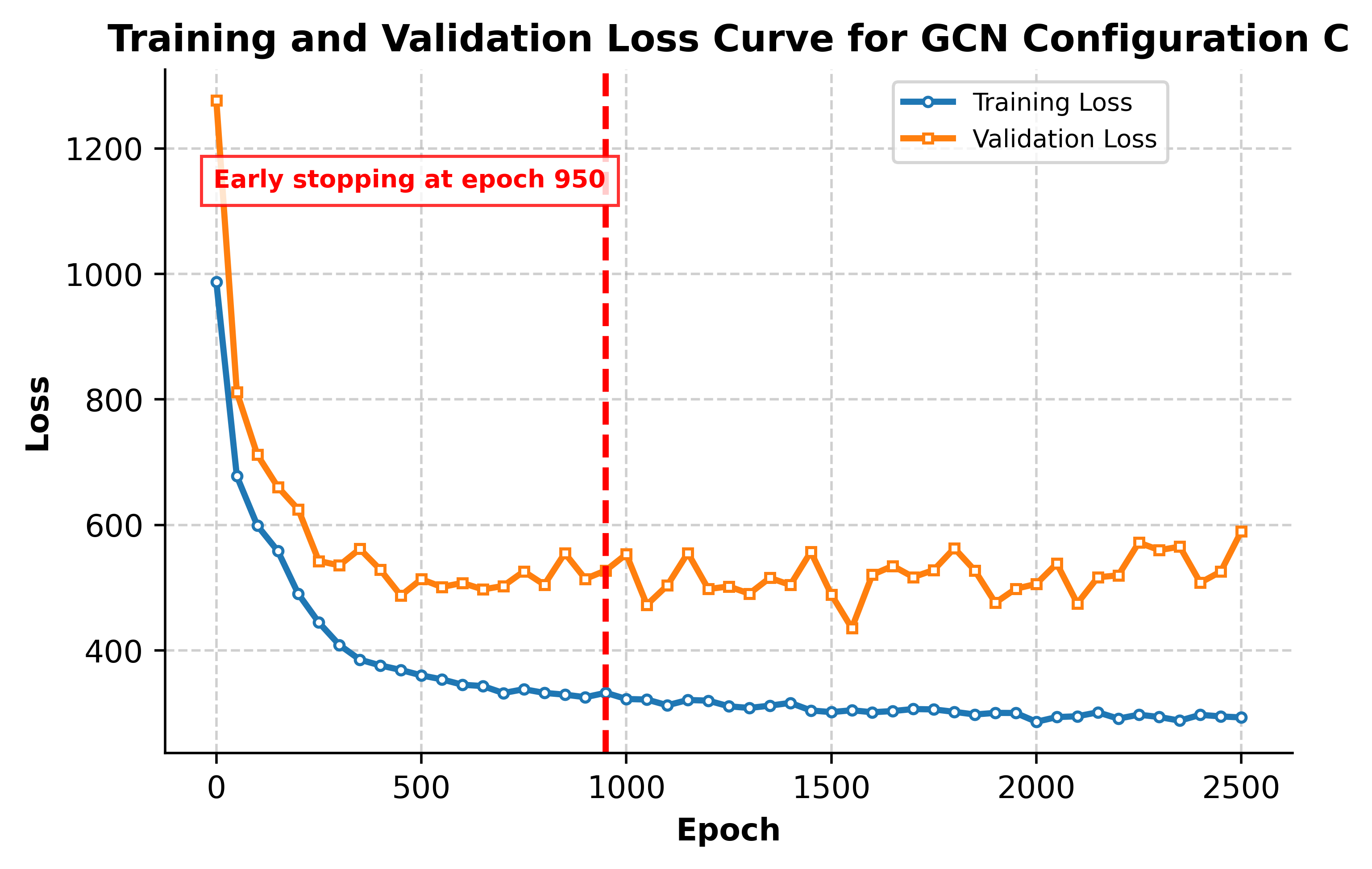}
        \subcaption{GCN Configuration C}
    \end{minipage}

    \vspace{2pt}
    
    \begin{minipage}{0.31\textwidth}
        \centering
        \includegraphics[width=\textwidth]{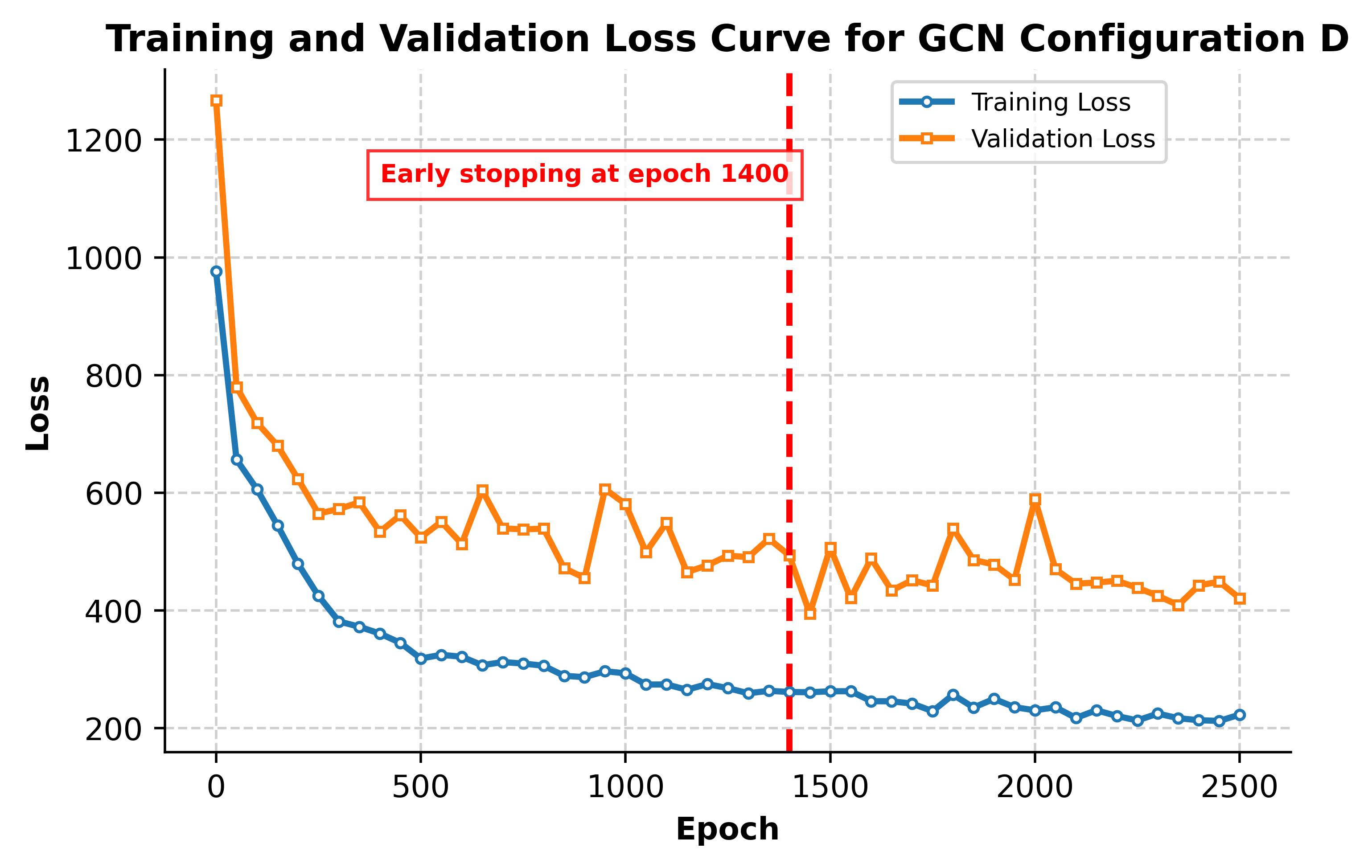}
        \subcaption{GCN Configuration D}
    \end{minipage}    
    \begin{minipage}{0.31\textwidth}
        \centering
        \includegraphics[width=\textwidth]{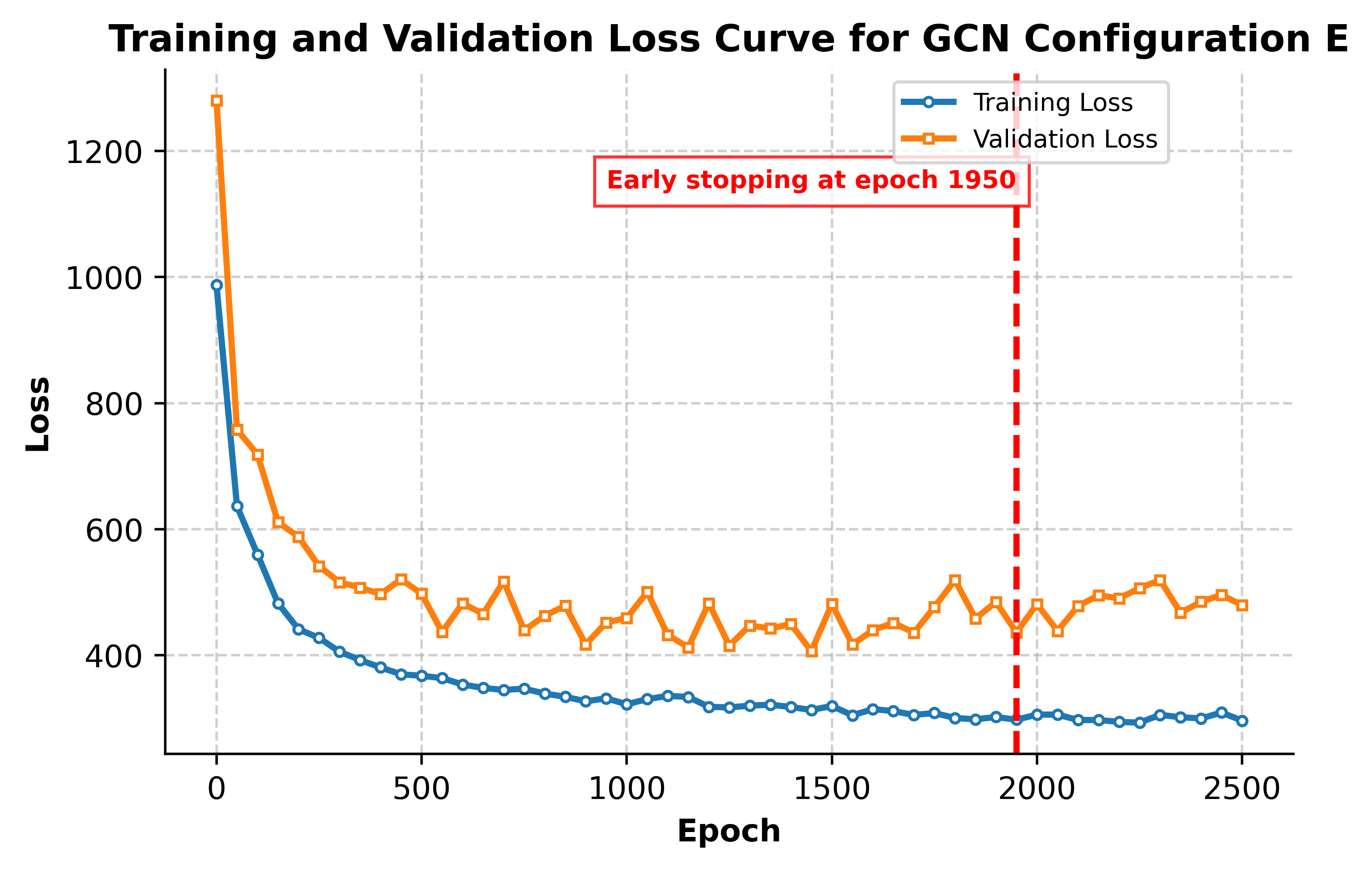}
        \subcaption{GCN Configuration E}
    \end{minipage}
    \begin{minipage}{0.31\textwidth}
        \centering
        \includegraphics[width=\textwidth]{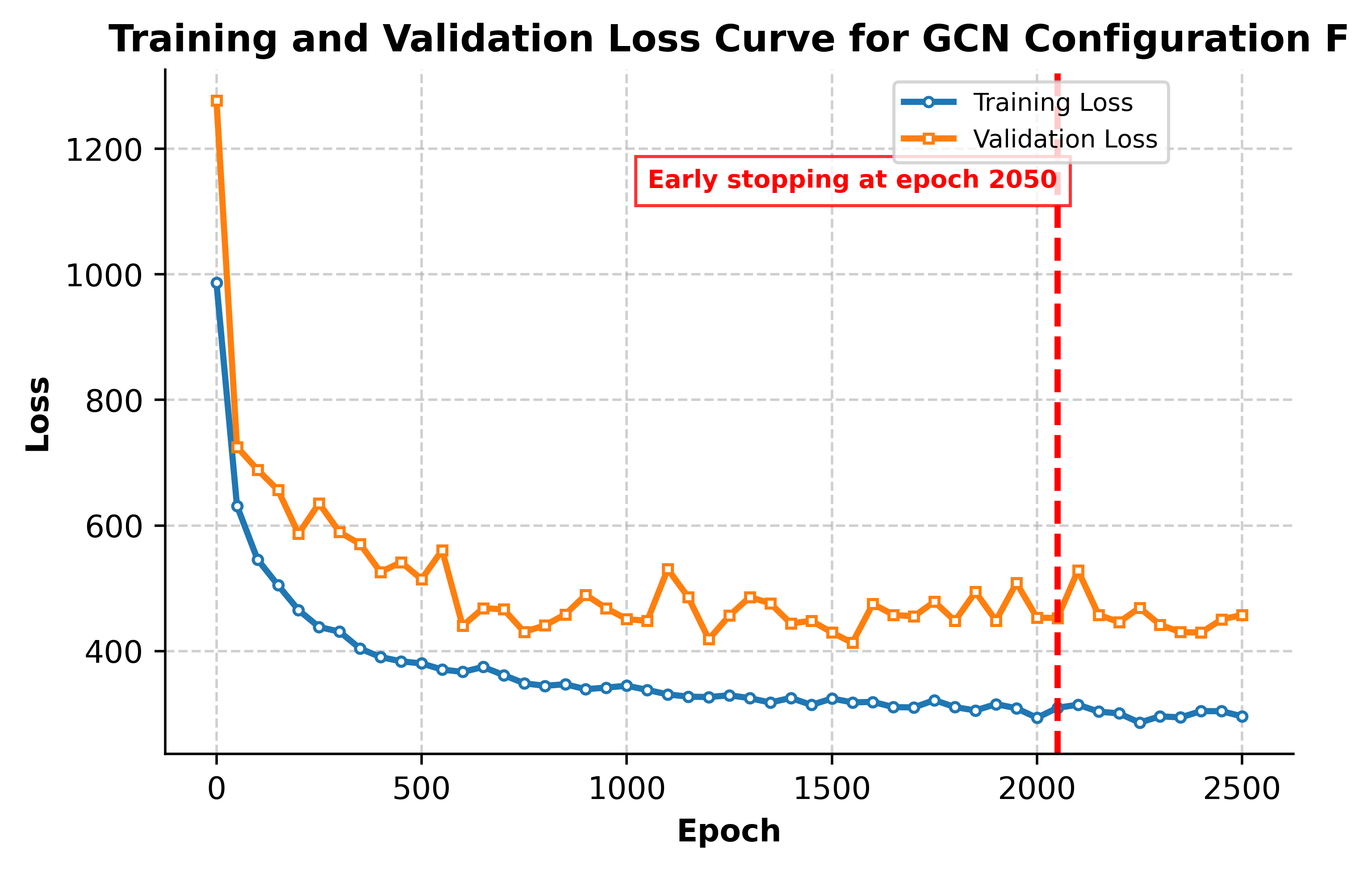}
        \subcaption{GCN Configuration F}
    \end{minipage}

    \vspace{2pt}

    \begin{minipage}{0.31\textwidth}
        \centering
        \includegraphics[width=\textwidth]{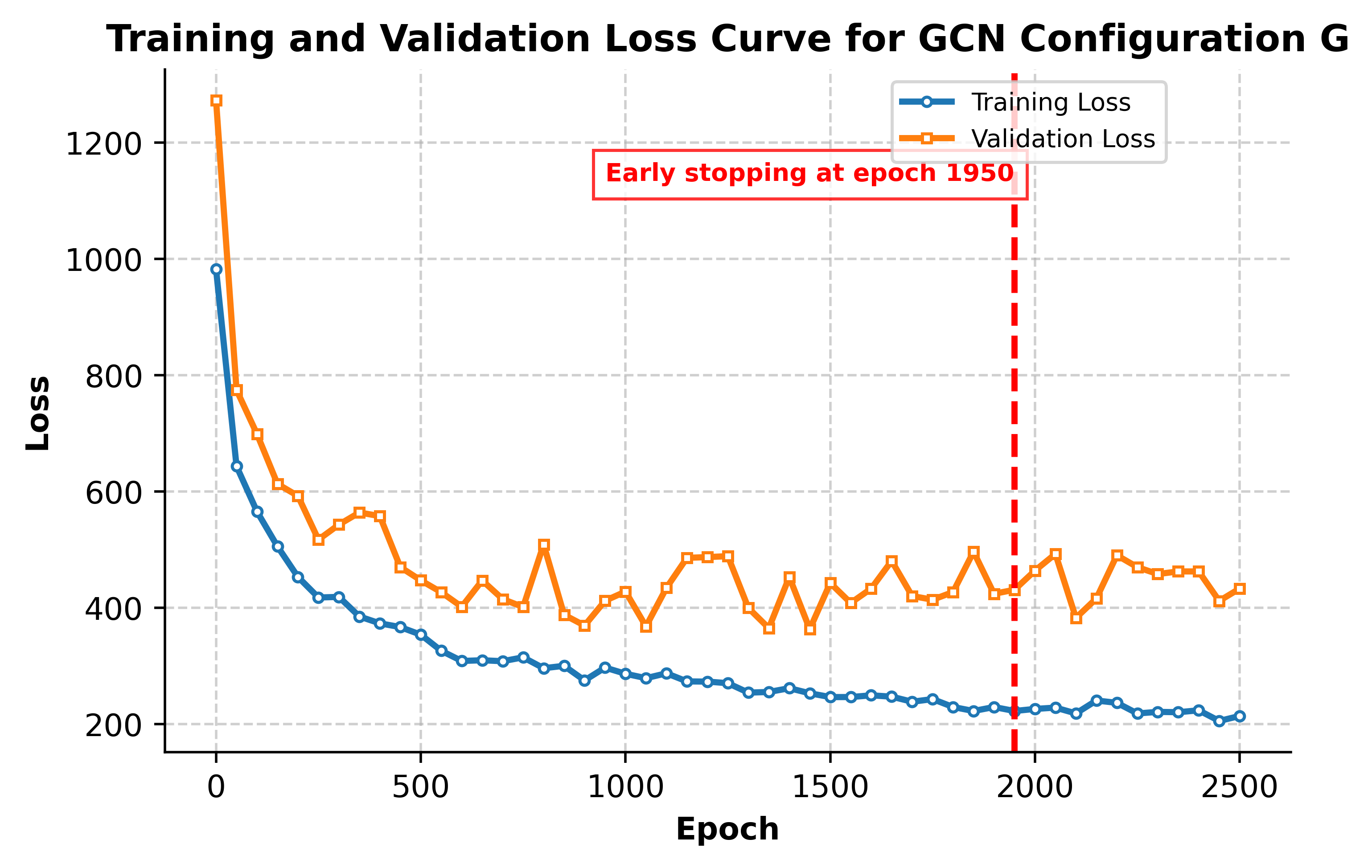}
        \subcaption{GCN Configuration G}
    \end{minipage}
    \begin{minipage}{0.31\textwidth}
        \centering
        \includegraphics[width=\textwidth]{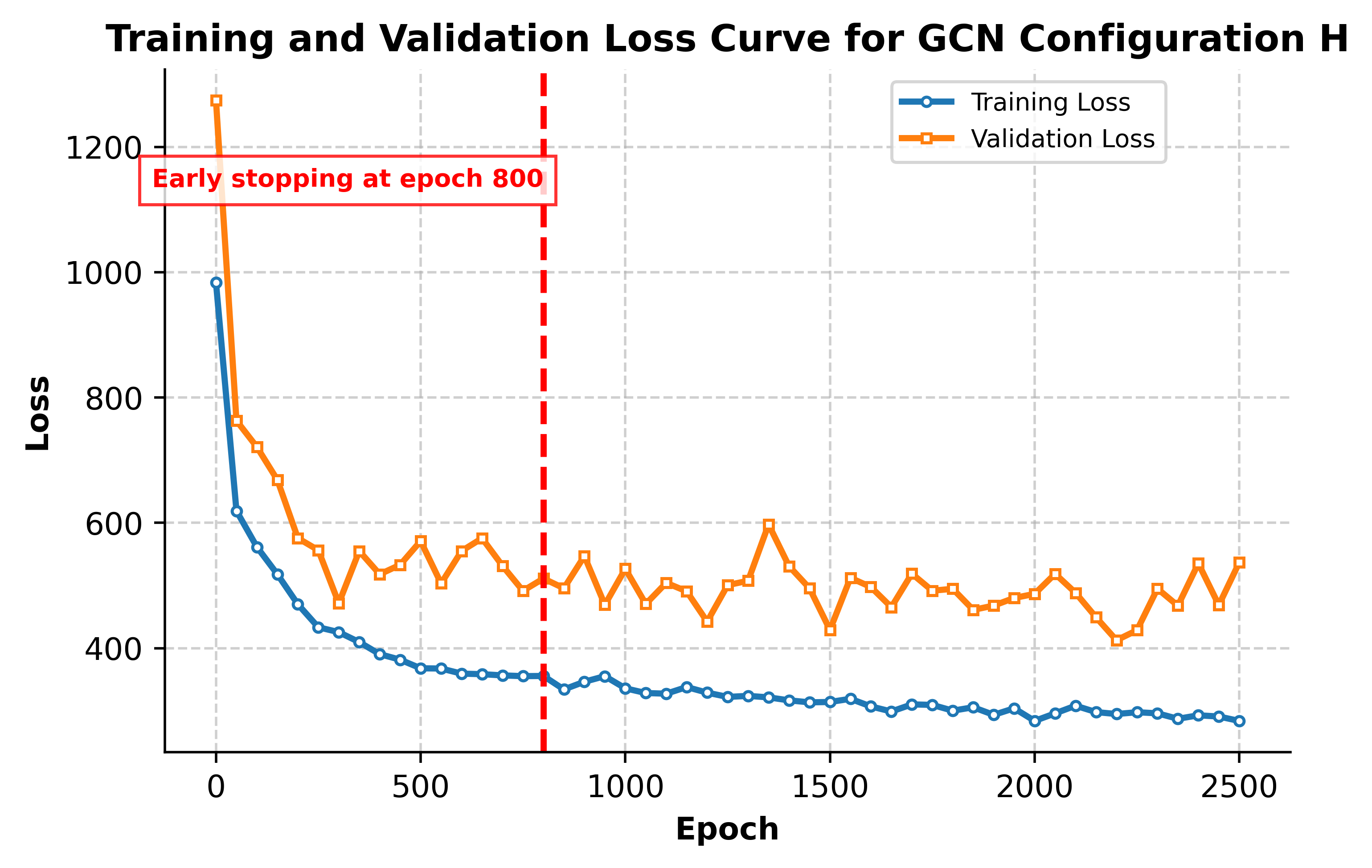}
        \subcaption{GCN Configuration H}
    \end{minipage}
    \begin{minipage}{0.31\textwidth}
        \centering
        \includegraphics[width=\textwidth]{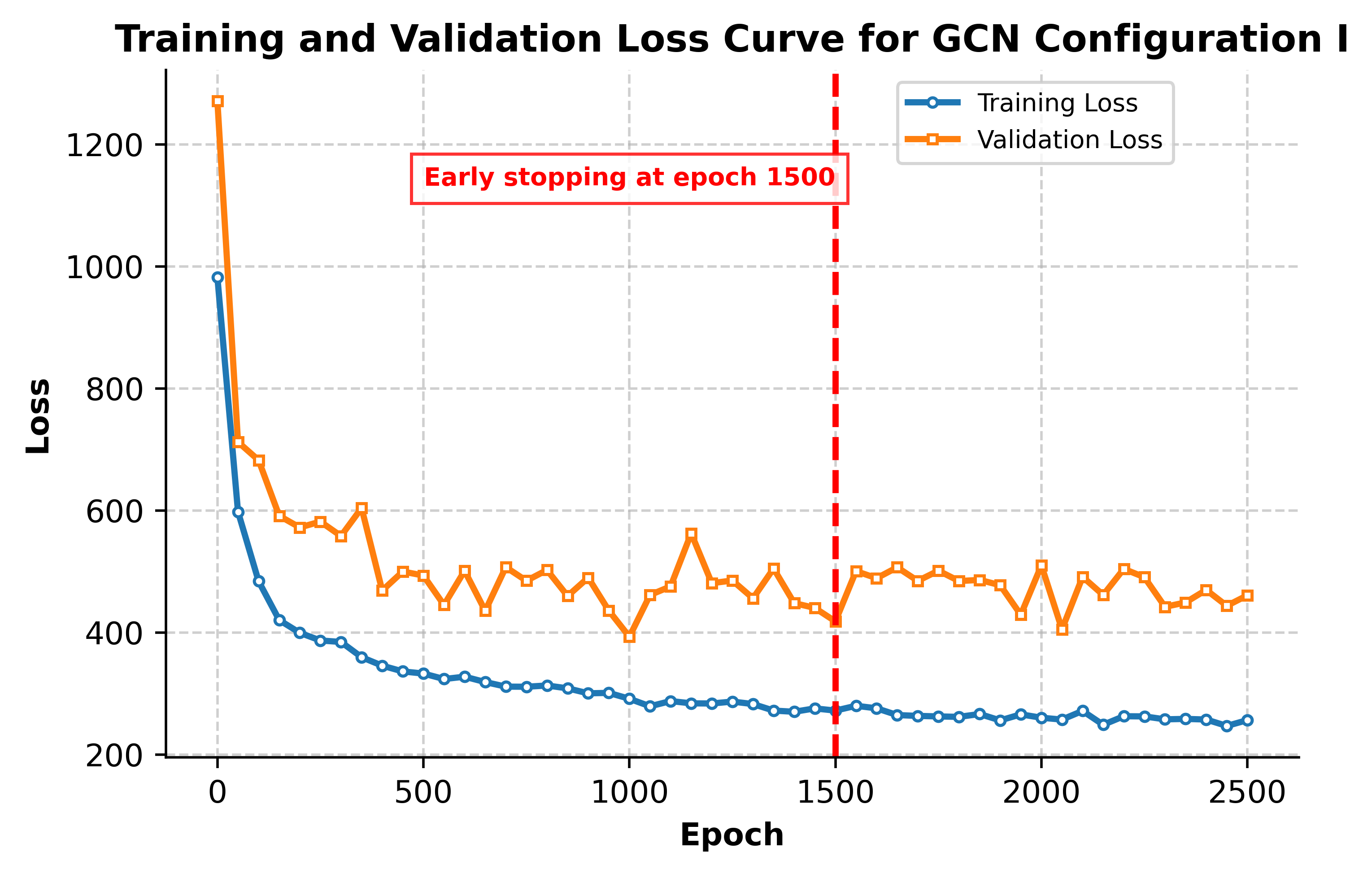}
        \subcaption{GCN Configuration I}
    \end{minipage} 

    \vspace{2pt}
    
    \begin{minipage}{0.31\textwidth}
        \centering
        \includegraphics[width=\textwidth]{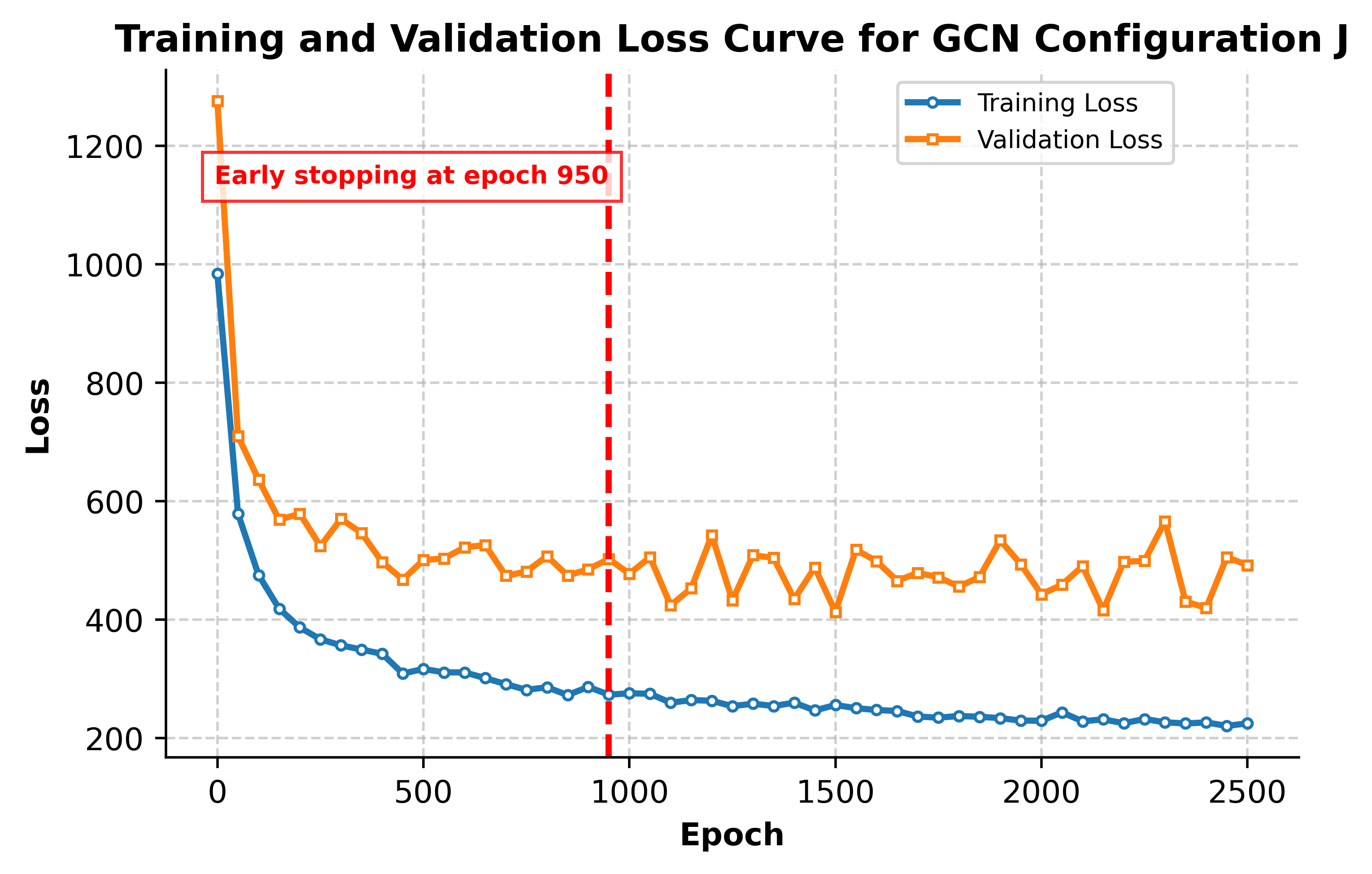}
        \subcaption{GCN Configuration J}
    \end{minipage}
    
    \caption[]{Loss curves for GCN configurations (A-J)}
    \label{fig:different_configuration_model_performance}
\end{figure}

\autoref{fig:different_configuration_model_performance} displays the training and validation loss curves for each of the GCN configurations (A-J) evaluated in this study. These curves provide a comprehensive overview of the model's learning behavior over 2500 epochs, illustrating how each configuration adapts to the training data while managing the balance between underfitting and overfitting. By examining these loss curves, we can assess the effectiveness of each configuration in terms of convergence speed, stability, and generalization ability.

The red vertical dashed lines in the plots represent the epoch where early stopping was activated, based on the validation loss. This mechanism is crucial for preventing overfitting, as it terminates training when further iterations no longer improve validation performance. 
The blue curve represents the training loss, while the orange curve depicts the validation loss. The gap between these two curves indicates the model's ability to generalize beyond the training data. A small, stable gap suggests that the model is not overfitting, while a large or widening gap may indicate that the model is memorizing the training data rather than learning the underlying patterns. 
This visualization helps us understand how each configuration performs under different settings, providing insight into why certain configurations, such as G, were more successful in balancing model complexity and generalization.

The purpose of including these curves is to offer a detailed analysis of the learning dynamics of each configuration. By visualizing the performance at each stage of the training process, we can better interpret why Configuration G was chosen as the optimal configuration for this task. It demonstrates early convergence and maintains a low validation loss, indicating its effectiveness in predicting link-level bicycling volumes with minimal risk of overfitting.

\subsection{Training and Validation Loss Curve for Model training at different sparsity levels}

\begin{figure}[H]
    \centering
    \begin{minipage}{0.31\textwidth}
        \centering
        \includegraphics[width=\textwidth]{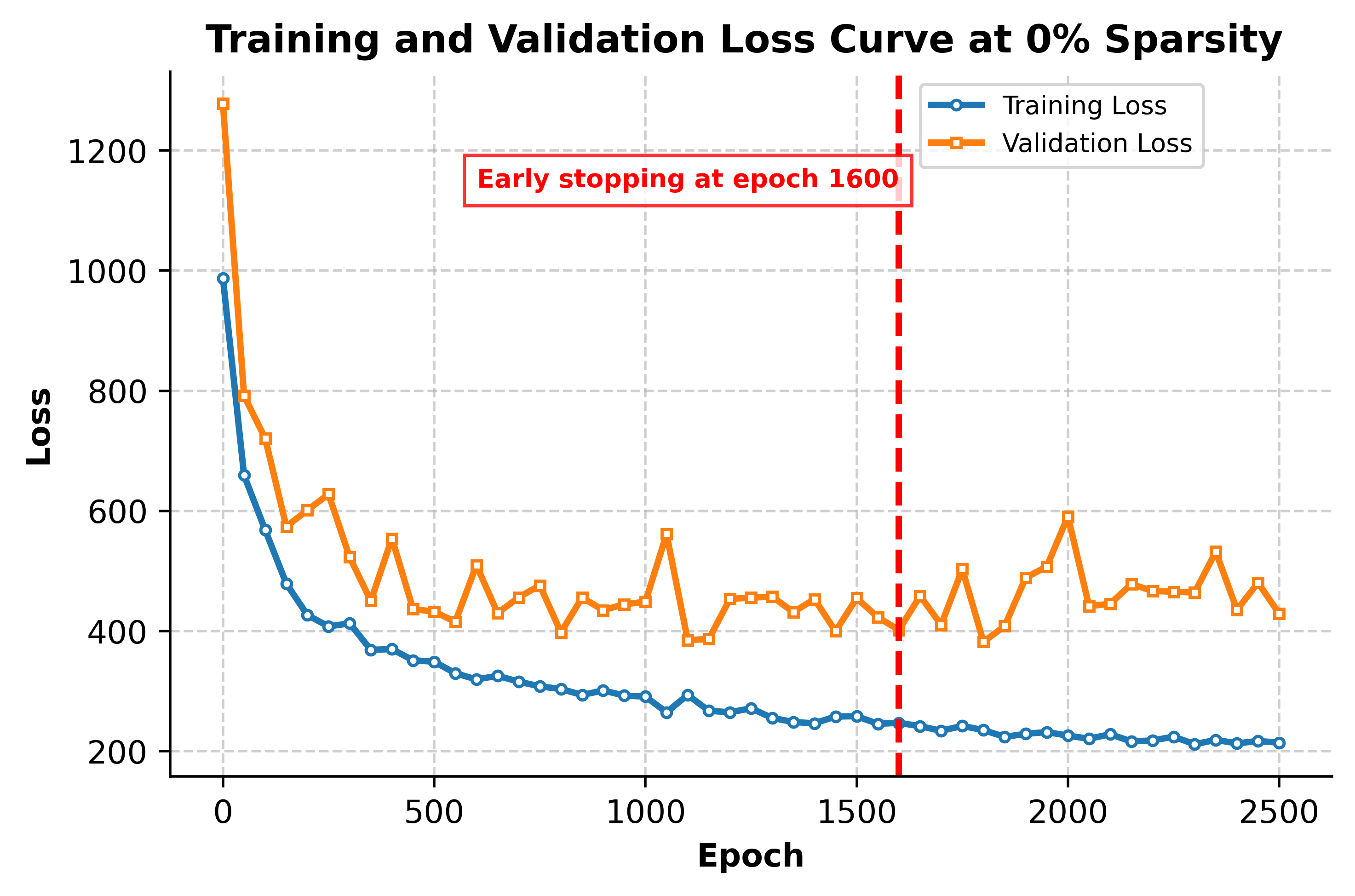}
    \end{minipage}
    \begin{minipage}{0.31\textwidth}
        \centering
        \includegraphics[width=\textwidth]{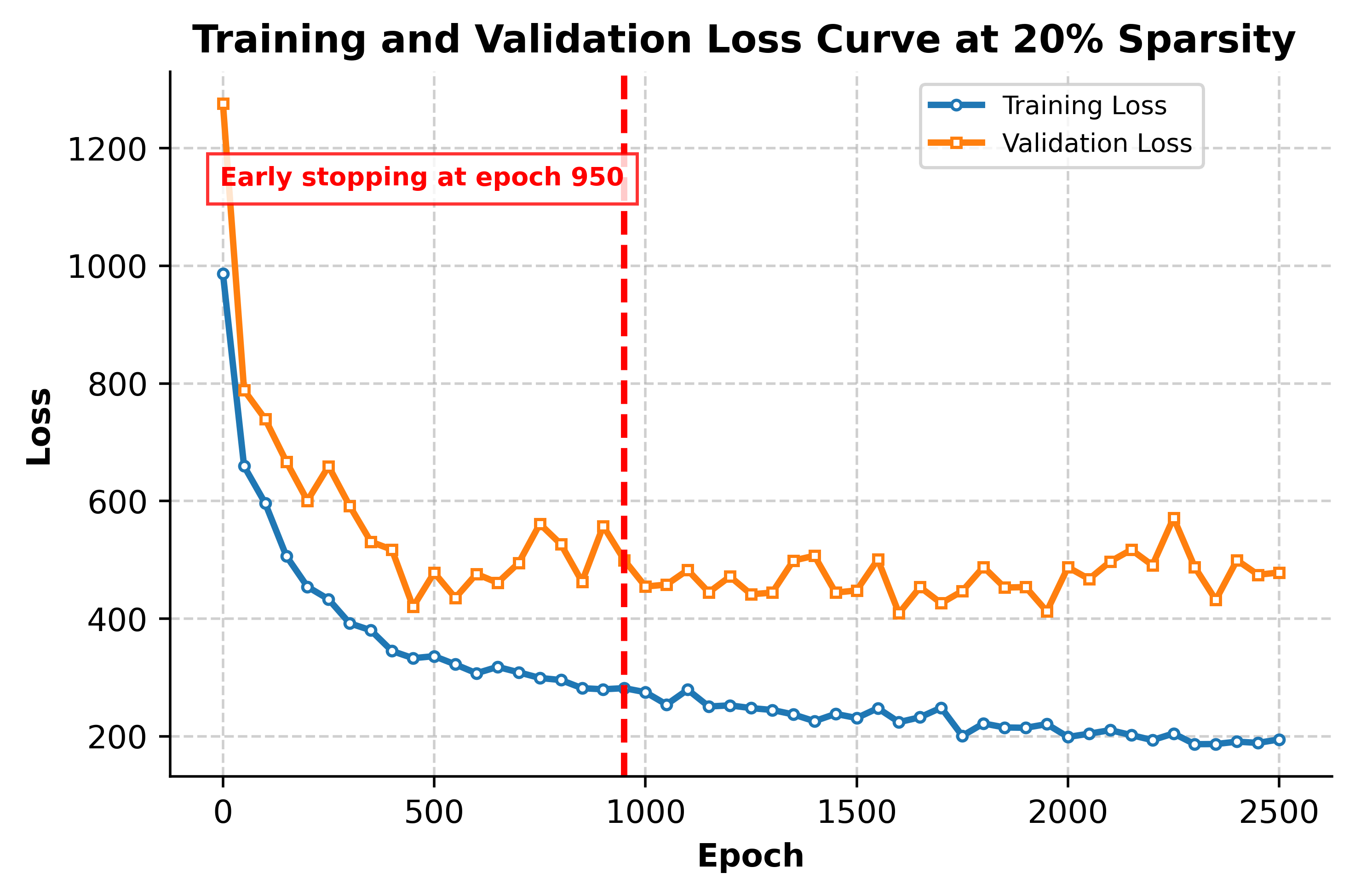}
    \end{minipage}
    \begin{minipage}{0.31\textwidth}
        \centering
        \includegraphics[width=\textwidth]{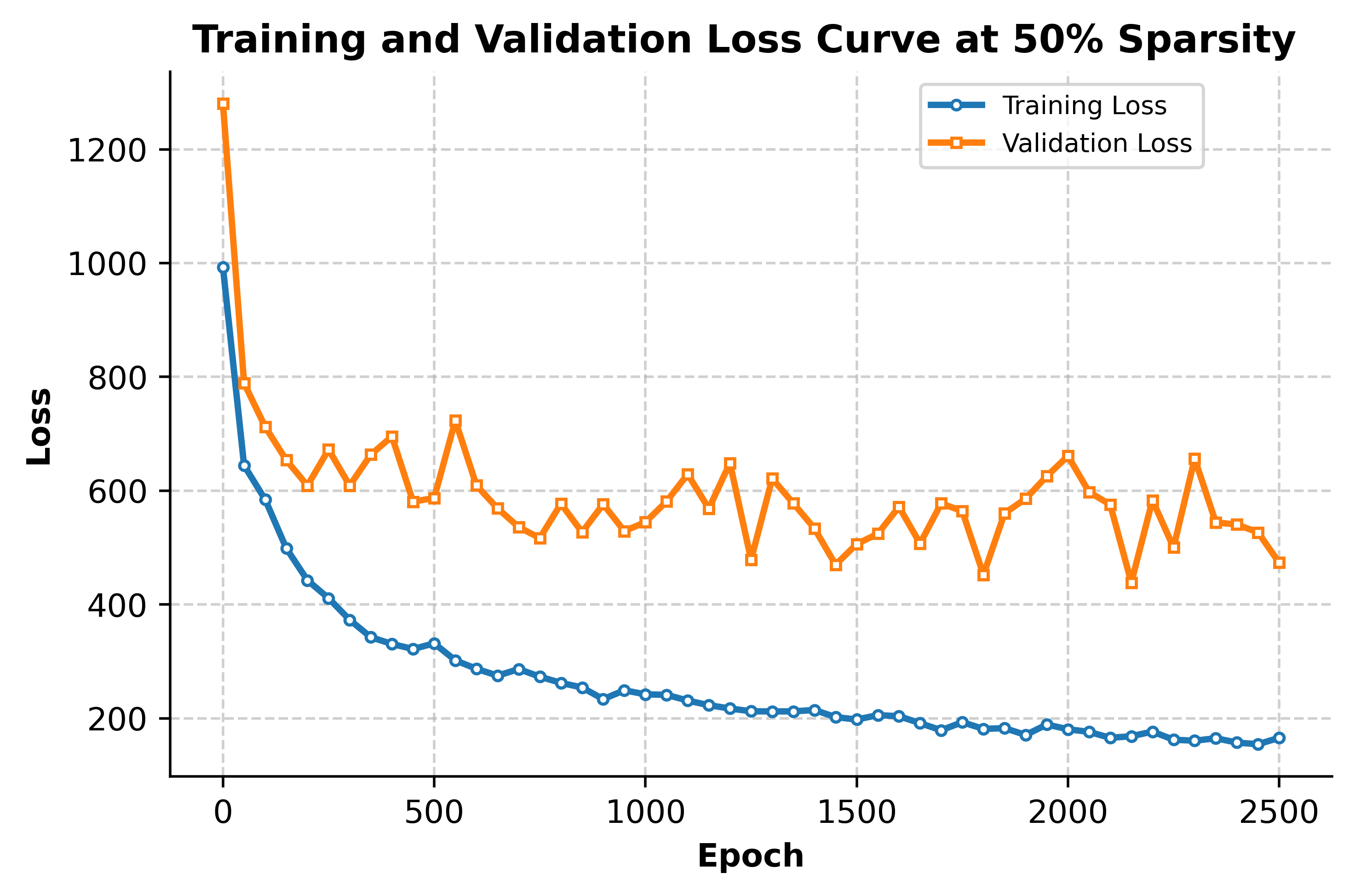}
    \end{minipage}

    \vspace{2pt}
    
    \begin{minipage}{0.31\textwidth}
        \centering
        \includegraphics[width=\textwidth]{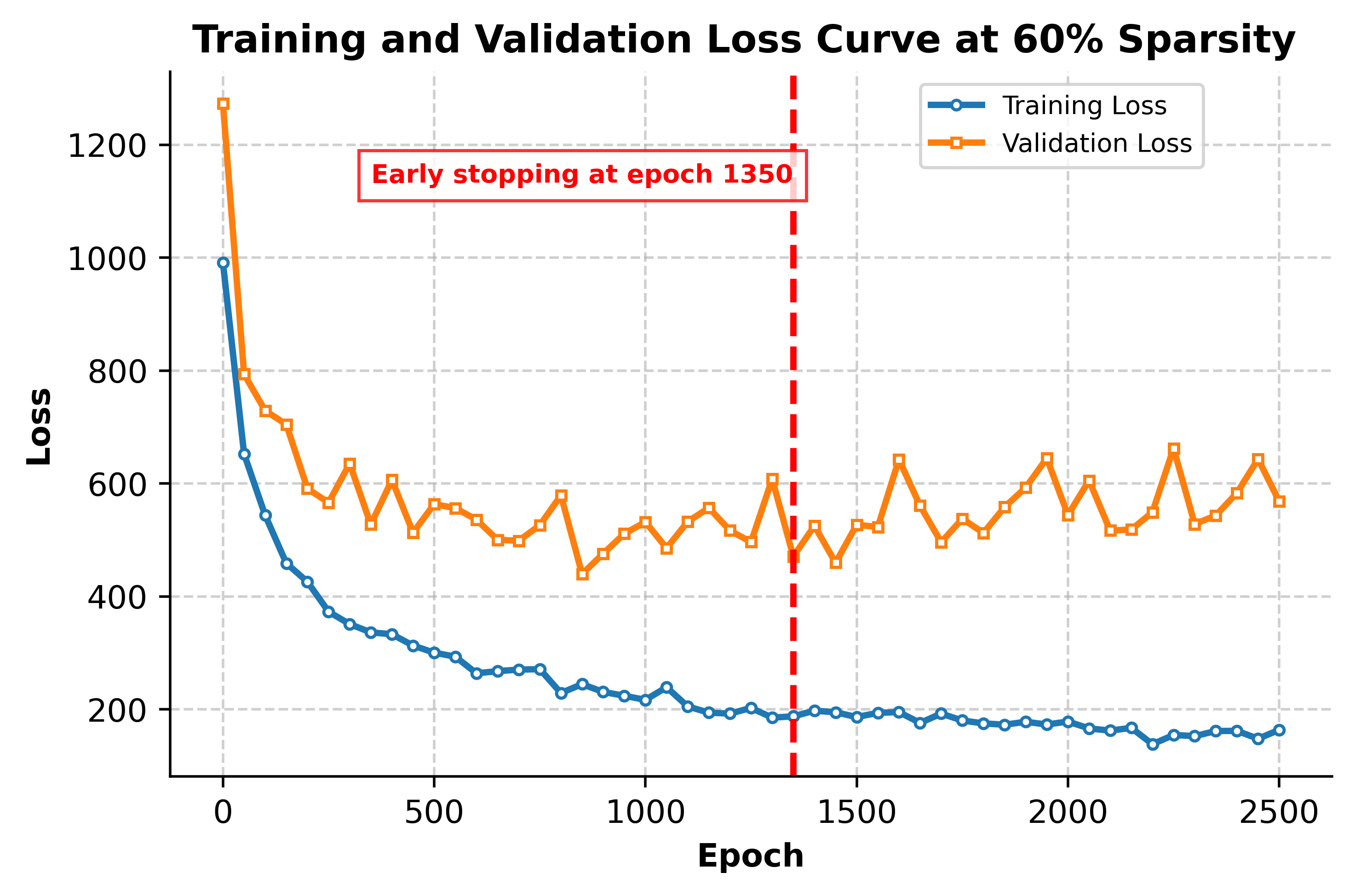}
    \end{minipage}
    \begin{minipage}{0.31\textwidth}
        \centering
        \includegraphics[width=\textwidth]{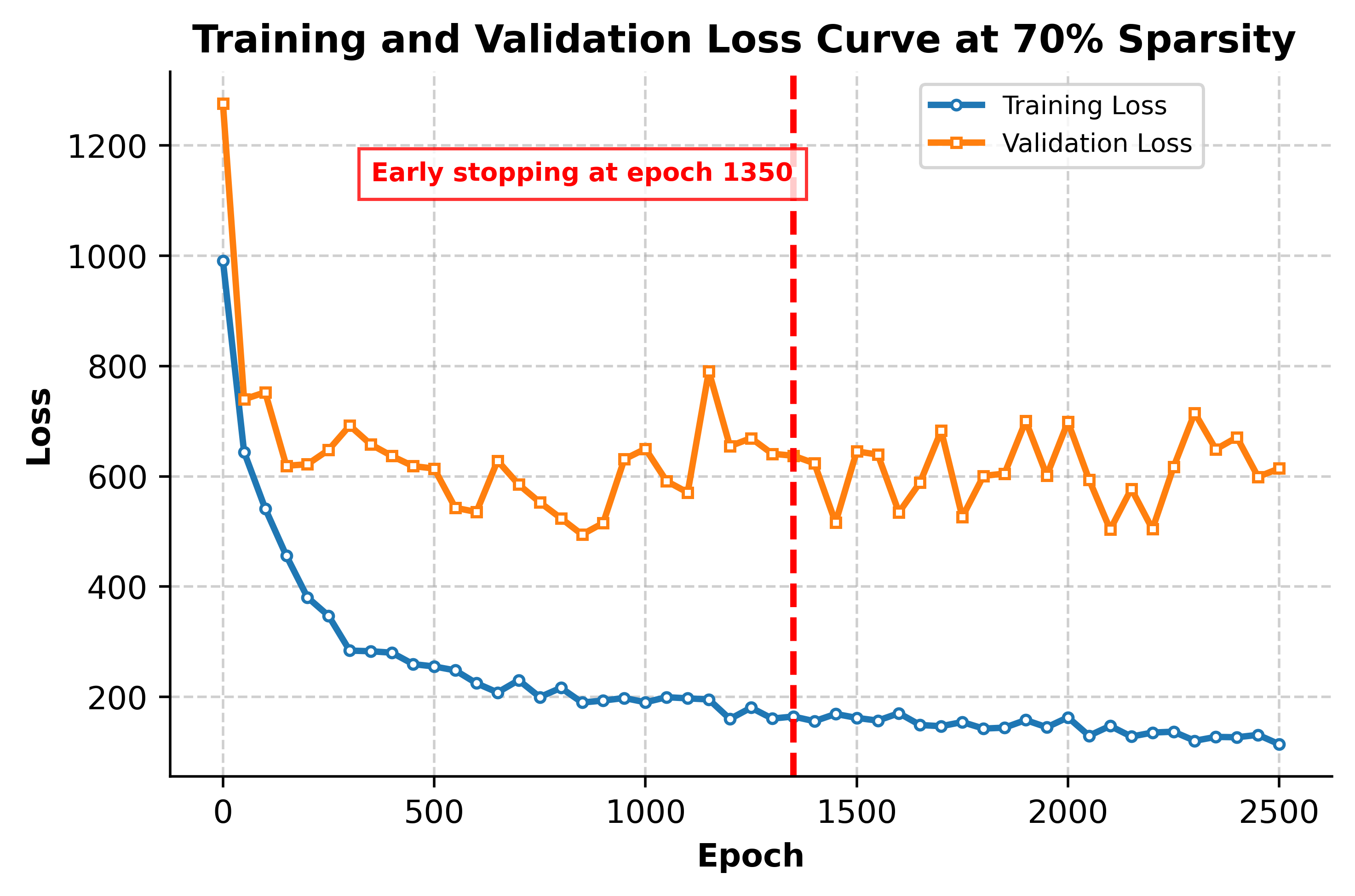}
    \end{minipage}
    \begin{minipage}{0.31\textwidth}
        \centering
        \includegraphics[width=\textwidth]{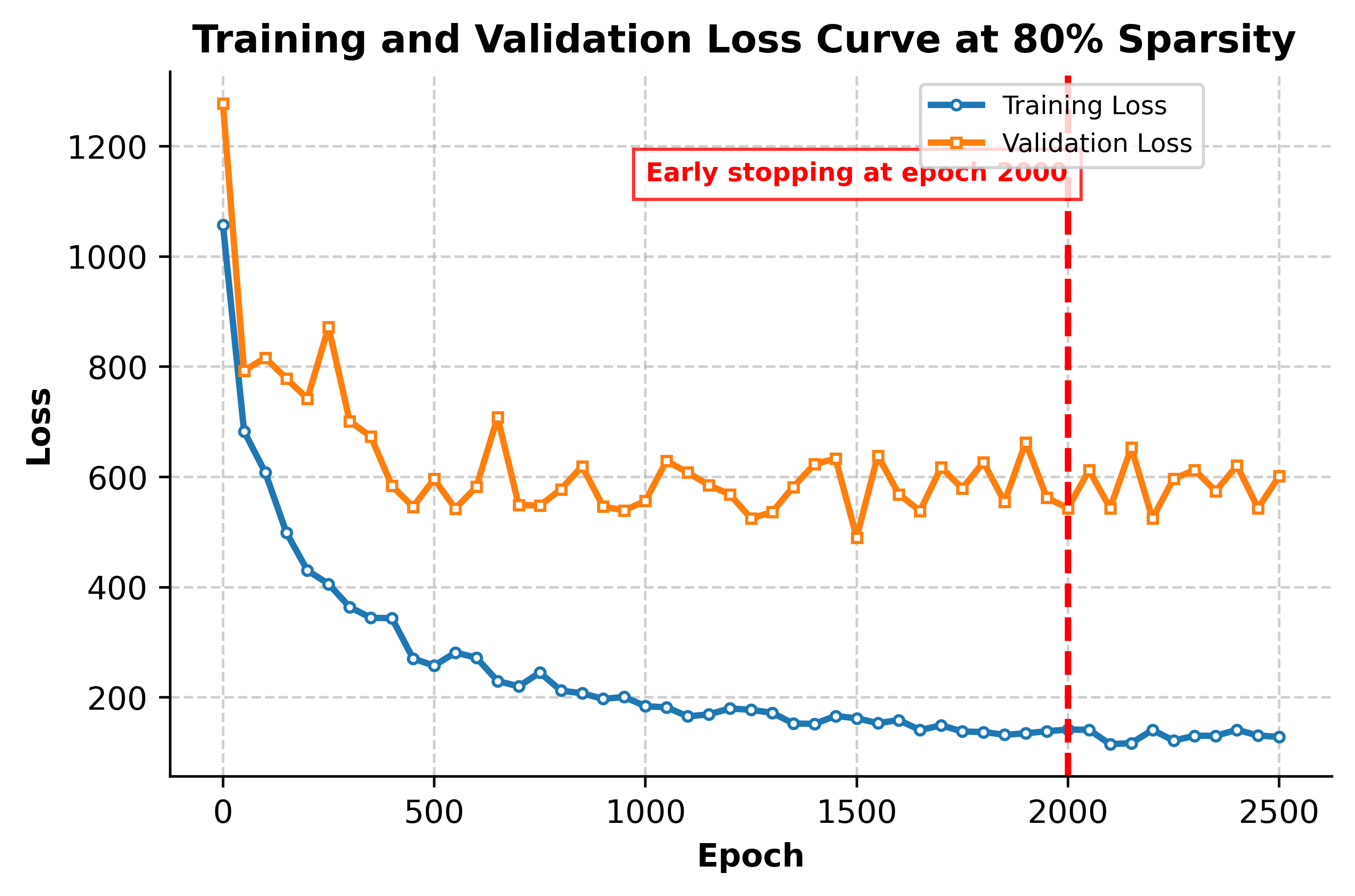}
    \end{minipage}

    \vspace{2pt} 

    \begin{minipage}{0.31\textwidth}
        \centering
        \includegraphics[width=\textwidth]{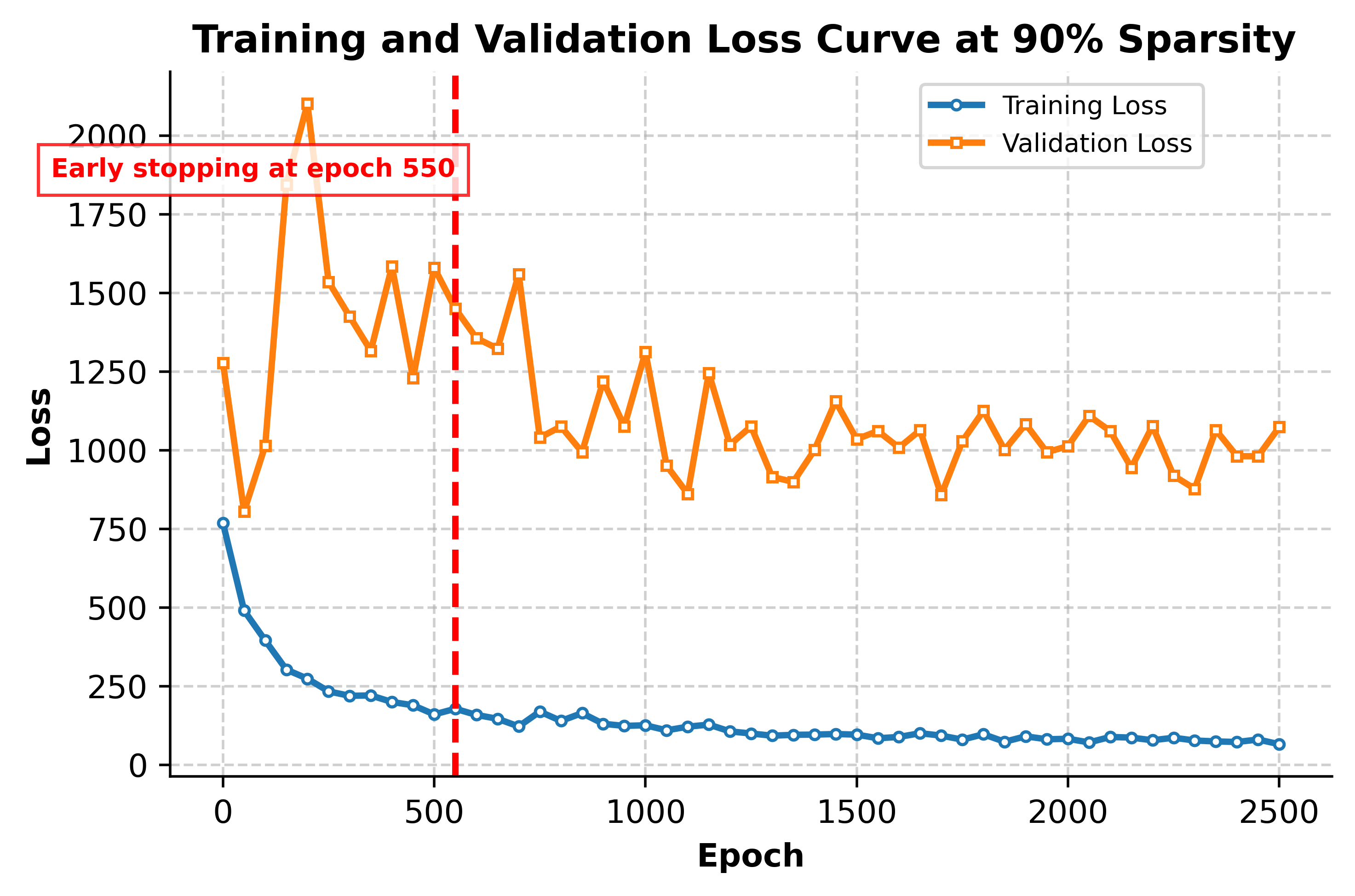}
    \end{minipage}
    \begin{minipage}{0.31\textwidth}
        \centering
        \includegraphics[width=\textwidth]{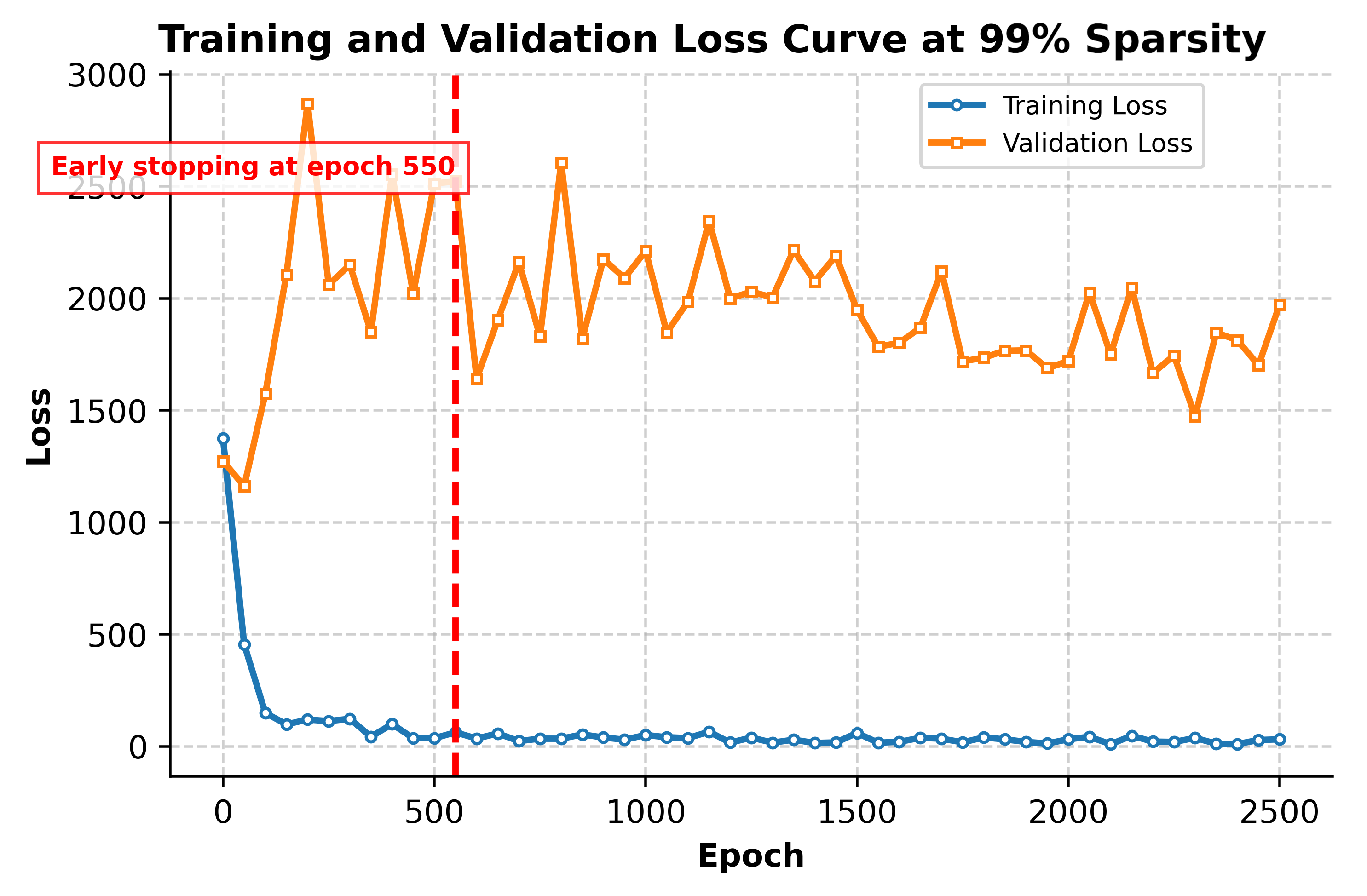}
    \end{minipage}

    \caption{\small Loss curves for GCN under different levels of sparsity}
    \label{fig:sparsity_level_model_performance}
\end{figure}

\autoref{fig:sparsity_level_model_performance} presents the training and validation loss curves for the GCN model (configuration G) across various levels of data sparsity, ranging from 0\% to 99\%. These plots are essential for understanding how the GCN model handles increasing data sparsity, which mirrors real-world scenarios where count data is extremely sparse.

As data sparsity increases, the model's learning process becomes more challenging. The earlier onset of early stopping, as seen at higher sparsity levels like 90\% and 99\%, suggests that the model struggles to find sufficient signal from the limited data available. The validation loss curves also show increased volatility as sparsity increases, indicating a difficulty in maintaining generalization.

At lower sparsity levels (e.g., 0\%, 20\%), the loss curves display smoother convergence, with the model able to train for longer before early stopping is triggered. In contrast, at high sparsity levels, the validation loss plateaus early, and the model starts to overfit as evidenced by the increasing gap between training and validation losses. This behavior highlights the model's sensitivity to sparsity, where insufficient data impairs its ability to generalize.

These loss curves provide crucial insight into how GCN models respond to data limitations, and serve as a valuable reference for future work aimed at improving model robustness in sparse data scenarios.

\clearpage 

\end{document}